\documentclass[pdflatex,sn-mathphys-num]{sn-jnl}

\usepackage{graphicx}%
\usepackage{multirow}%
\usepackage{amsmath,amssymb,amsfonts}%
\usepackage{amsthm}%
\usepackage{mathrsfs}%
\usepackage[title]{appendix}%
\usepackage{xcolor}%
\usepackage{textcomp}%
\usepackage{manyfoot}%
\usepackage{booktabs}%
\usepackage{algorithm}%
\usepackage{algorithmicx}%
\usepackage{algpseudocode}%
\usepackage{listings}%
\usepackage{float}
\usepackage{hyperref}
\usepackage{color}
\usepackage{soul}

\usepackage[pagewise]{lineno}

\theoremstyle{thmstyleone}%
\theoremstyle{thmstyletwo}%

\theoremstyle{thmstylethree}%

\renewcommand{\figurename}{Figure}

\begin{document}

\title[Through-Foliage Surface-Temperature Reconstruction for early Wildfire Detection]{Through-Foliage Surface Temperature Reconstruction for Early Wildfire Detection}


\author[1]{\fnm{Mohamed} \sur{Youssef}}\email{mohamed.youssef@jku.at} \equalcont{These authors contributed equally to this work.}

\author[1]{\fnm{Lukas} \sur{Brunner}}\email{lukas.brunner@profactor.at} \equalcont{These authors contributed equally to this work.}

\author[2]{\fnm{Klaus} \sur{Rundhammer}}\email{k.rundhammer@ffagatha.at}

\author[3]{\fnm{Gerald} \sur{Czech}}\email{gerald.czech@ooelfv.at}

\author*[1]{\fnm{Oliver} \sur{Bimber}}\email{oliver.bimber@jku.at} \equalcont{These authors contributed equally to this work.}

\affil[1]{\orgdiv{Department of Computer Science}, \orgname{Johannes Kepler University}, \orgaddress{\street{Altenbergerstr. 65}, \city{Linz}, \postcode{4040}, \country{Austria}}}

\affil[2]{\orgname{Fire Brigade St. Agatha }, \orgaddress{\street{St. Agatha 19}, \city{Bad Goisern}, \postcode{4822}, \country{Austria}}}

\affil[3]{\orgname{Upper Austria Fire Brigade Headquarter}, \orgaddress{\street{Petzoldstraße 43}, \city{Linz}, \postcode{4021}, \country{Austria}}}

\abstract{We present a method to reconstruct surface temperatures through forest vegetation by combining signal processing and machine learning, enabling fully automated aerial wildfire monitoring with drones for early fire detection. Synthetic aperture (SA) sensing reduces canopy occlusion but introduces thermal blur. To overcome this, we train a visual state space model to recover subtle thermal signals of partially occluded soil and fire hotspots from blurred data. To address limited real-world training data, we generate realistic surface temperature simulations using a latent diffusion model, temperature augmentation, and procedural thermal forest modeling. On simulated datasets, our method reduces RMSE by 2-2.5 versus conventional thermal and uncorrected SA imaging; in field experiments on hotspots, RMSE improved by 12.8-fold and 2.6-fold, respectively. Our approach also generalizes to other thermal signals, including human signatures, capturing morphology and extent --critical where simple thresholding fails-- while conventional imaging struggles with partial occlusion.
}




\maketitle

\section{Introduction}\label{Sec:Introduction}
Recent wildfire events demonstrate that their frequency and scale are rising dramatically due to climate change. These unplanned and uncontrolled disasters devastate natural habitats, with severe consequences for both local and global ecosystems \cite{jolly2015climate}. Consequently, the early detection of ground and surface fires while they are still small and manageable is critical for minimizing ecological damage, public risk, and economic costs \cite{pradhan2007forest}. Modern detection systems -- including terrestrial networks (e.g., fixed cameras on watchtowers) \cite{hough2007vision, lin2018fuzzy}, satellites \cite{he2012enhancement, csiszar2014active}, and aerial sensors (e.g., cameras or smoke detectors on UAVs or manned aircraft) \cite{krull2012early, von2010integrated, jiao2019deep, bradley2011georeferenced, wardihani2018real, kanand2020wildfire, dimitropoulos2012flame, celik2010fast, avgerinakis2012smoke, gunay2010fire, mueller2013optical, zhang2014improved, ghamry2016cooperative, martinez2011automatic, ambrosia2011ikhana, martinez2005fire, valero2018use, valero2018automated, yuan2017fire, srinivas2019fog, lee2017deep, novac2020framework, alexandrov2019analysis, yadav2020deep, barmpoutis2020early, zhao2018saliency, cheng2019smoke, shamsoshoara2021aerial, huseynov2007infrared, deng2019fire, ciprian2023fire, seydi2022fire, benzekri2020early, barmpoutis2020review, bouguettaya2022review, gaur2020video, allison2016airborne, sousa2019classification, viseras2019wildfire, lin2021autonomous, wright2004real} --all share a significant limitation: occlusion by vegetation, which prevents the early detection of wildfires.

Terrestrial systems, which employ either individual sensors (fixed, PTZ, or 360-degree cameras \cite{hough2007vision}) or networks of ground sensors \cite{lin2018fuzzy}, are typically mounted on watchtowers. Their static nature limits scalability, and they are hampered by maintenance challenges and the need for independent power sources \cite{lin2018fuzzy}. Furthermore, these systems are often destroyed during a fire, incurring additional replacement costs. Satellite imagery provides broad coverage and is a vital tool for strategic fire management, capable of identifying medium to large fires using automated algorithms \cite{he2012enhancement, csiszar2014active}. However, its utility for early detection is constrained by low temporal resolution, high deployment costs, inflexibility, and a lack of real-time data, making the timely identification of small fire hotspots difficult \cite{martinez2008computer}. Unmanned Aerial Vehicles (UAVs) offer a promising alternative due to their maneuverability, autonomy, relatively low cost, and advanced computational capabilities, making them well-suited for remote sensing in hostile environments \cite{nex2019preface}. Equipped with a variety of sensors --including visual \cite{jiao2019deep, martinez2011automatic}, infrared \cite{bradley2011georeferenced}, thermal \cite{wardihani2018real, kanand2020wildfire}, and gas or smoke detectors \cite{krull2012early, von2010integrated}, UAVs have been widely deployed for fire monitoring. These systems leverage the distinct signatures of fire, such as heat, smoke, flickering motion, and specific bio-products. Detection methods using visual or thermal data often employ classical computer vision \cite{dimitropoulos2012flame, celik2010fast, avgerinakis2012smoke, gunay2010fire, mueller2013optical, zhang2014improved} and machine learning techniques \cite{ ghamry2016cooperative, martinez2011automatic, ambrosia2011ikhana, martinez2005fire, valero2018use, valero2018automated, yuan2017fire, srinivas2019fog, lee2017deep, novac2020framework, alexandrov2019analysis, yadav2020deep, barmpoutis2020early, zhao2018saliency, cheng2019smoke, shamsoshoara2021aerial, huseynov2007infrared, deng2019fire}. Traditional computer vision approaches rely on features like color thresholds \cite{dimitropoulos2012flame, celik2010fast}, spatial and texture characteristics \cite{avgerinakis2012smoke, mueller2013optical}, temporal pixel variation \cite{gunay2010fire}, and motion in image sequences \cite{zhang2014improved} to identify fire, smoke, and model fire spread \cite{ghamry2016cooperative} or volume \cite{martinez2011automatic}. Similarly, early thermal image analysis uses thresholding \cite{martinez2005fire, sousa2019classification, viseras2019wildfire, lin2021autonomous, wright2004real}, contour detection \cite{valero2018use}, segmentation \cite{valero2018automated}, and spatiotemporal analysis \cite{yuan2017fire} to locate wildfires \cite{martinez2005fire, valero2018use, valero2018automated, yuan2017fire} and hotspots \cite{viseras2019wildfire, lin2021autonomous, wright2004real}. More recent advances utilize deep learning for image classification (e.g., AlexNet \cite{srinivas2019fog}, GoogleNet \cite{lee2017deep}), object detection (e.g., R-CNN \cite{alexandrov2019analysis}, YOLO \cite{yadav2020deep, jiao2019deep}), and semantic segmentation (e.g., DeepLab \cite{zhao2018saliency, cheng2019smoke}, U-Net \cite{shamsoshoara2021aerial}). These techniques have also been applied directly to thermal imagery \cite{huseynov2007infrared, deng2019fire}. To enhance accuracy, some solutions fuse data from thermal, optical \cite{ciprian2023fire, seydi2022fire}, and other sensors \cite{benzekri2020early}. For comprehensive reviews of UAV-based wildfire detection, see \cite{barmpoutis2020review, bouguettaya2022review, gaur2020video, yuan2015survey}.

A critical shortcoming of current approaches --both traditional and deep learning-- is that they typically model fire and smoke as a simple "bag of features" (color, texture, shape). This makes them prone to misclassifications and false alarms, especially under conditions of limited visibility \cite{jiao2019deep, yadav2020deep, zhao2018saliency, shamsoshoara2021aerial, ciprian2023fire, benzekri2020early, yuan2015survey}. While thermal imaging can often penetrate smoke, the fundamental challenge remains occlusion by vegetation. Small, smoldering ground fires (the earliest stage of many wildfires) produce no visible flames or smoke. If these nascent hotspots are beneath a dense forest canopy, they become virtually undetectable. Their temperatures can be weak and a simple temperature threshold is often insufficient for effective detection, as it is easily invalidated by sunlight reflection and thermal radiation from the tree canopy and soil. The morphological characteristics, such as shape and extent, of heated areas are equally important for accurate classification. By the time significant smoke and flames are developed, however, it is often too late for an early and efficient response. 

Here we present a method for detecting occluded wildfires by leveraging a combination of signal processing and machine learning. Drones are used to
capture thermal images in a grid pattern over forested areas. These images are first
computationally combined using an optical synthetic aperture sensing technique called
Airborne Optical Sectioning (AOS) [51–70] to reduce occlusion caused by vegetation.
This suppresses the thermal signal from the forest canopy to enhance the surface temperature signal. However, the clarity of the resulting integral images remains limited
by vegetation density, which introduces persistent temperature blur and limits accurate recovery of surface temperatures. Whereas AOS previously has been used for occlusion removal that supports shape detection (e.g. person classification in search and rescue), here we instead significantly shift the paradigm  from visibility enhancement to radiometric reconstruction. 

To achieve this, we train a visual state space model for image restoration \cite{10843251} to reconstruct the extremely weak thermal signals of ground fires obscured by dense canopy, which are normally undetectable. Unlike conventional de-convolution approaches, which are inherently ill-posed and rely on accurate knowledge of the point spread function (PSF), our problem involves a PSF that arises from a complex combination of synthetic aperture effects and unknown occlusion-induced structures, making reliable inversion infeasible. Instead, the proposed method learns a direct radiometric mapping between biased AOS integral images and their corresponding ground-truth surface temperatures, enabling through-foliage reconstruction of both absolute temperature and fire morphology.

Training such a model requires extensive data on occluded ground and surface fires, which is not available in practice. Existing wildfire databases \cite{shamsoshoara2021aerial, hopkins2023flame, hopkins2024flame} primarily contain imagery of developed, visible fires. They are insufficient and too small for efficient training. Furthermore, acquiring pixel-accurate ground-truth surface temperatures beneath a forest canopy is fundamentally infeasible without physically removing the occluding vegetation. 

To address this data scarcity, we generated a large-scale synthetic dataset  through the following procedure. 
We began with a limited dataset consisting of several hundred thermal images of real, non-occluded ground and surface fires, captured during a high-altitude mountain wildfire above the treeline. These real observations were used to define the initial temperature distributions and morphological characteristics of fire patterns.
This initial dataset was significantly expanded to many thousands of samples by embedding latent diffusion \cite{rombach2022high} into a vector quantized variational autoencoder \cite{van2017neural}, and by additional temperature augmentation. By anchoring the generative process in real measurements, the synthesized data remain physically and statistically consistent with real  wildfire signatures. Subsequently, we superimposed occluding vegetation onto these samples using thermal procedural forest simulation, in which we modeled varying conditions such as local foliage density, ambient temperature, and solar azimuth. This process yielded hundreds of thousands of simulated thermal drone images of occluded ground and surface fires. Finally, we processed these images with AOS, enabling our model to learn the mapping between the AOS integral images (that suppress occlusion but leave thermal blur) and the corresponding occlusion-free ground truth.

Our trained model facilitates the reconstruction of surface temperatures through occluding dense forest canopy. On simulated data across a wide range of ambient temperatures, forest densities, and sunlight conditions, our method reduced RMSE by a factor of 2 to 2.5 compared to both conventional thermal imaging and uncorrected AOS integrals. The improvement was even more pronounced in field experiments focusing on high-temperature hotspots, where we observed a 12.8-fold RMSE gain over conventional methods and a 2.6-fold gain over uncorrected AOS. Crucially, our technique successfully revealed the complete morphological characteristics, such as shape and extent, of fire hotspots that conventional imaging misses due to occlusion. As our approach is highly scalable and real-time, it has the potential to enable fully automated aerial wildfire monitoring across large forested terrain using high-endurance autonomous drones. 

\section{Results}\label{Sec:Results}
In the following we validate our method through simulations and field experiments, also showing the model's ability to generalize to other signals, such as human thermal signatures relevant for search and rescue missions.

We first detail the procedure for generating the training data. This involves creating realistic forest soil temperatures by latent diffusion and vector-quantized variational autoencoder, expanding a limited set of aerial thermal images of real surface fires, and overlaying them with thermal radiation from a simulated, procedurally generated forest. This process yields an extensive database of synthetic aerial thermal imagery for training and validation, depicting soil temperatures (including surface fires) obscured by forest vegetation.

Next, we describe the application of synthetic aperture sensing (Airborne Optical Sectioning, AOS) to suppress occlusions caused by forest vegetation -- a technique applicable to both real and simulated images. We then explain how a visual state space model for temperature restoration is trained and validated on the generated dataset. This model is designed to recover accurate surface temperatures from the faint or blurred temperature signals in AOS integral images. Finally, we present a comprehensive validation of our approach through quantitative metrics and visual results from both simulated data and field experiments.

\subsection{Generating Surface Temperatures}\label{Sec:GeneratingSurfaceTemperatures}
\begin{figure}[H]
\centering
\includegraphics[width=1.0\textwidth]{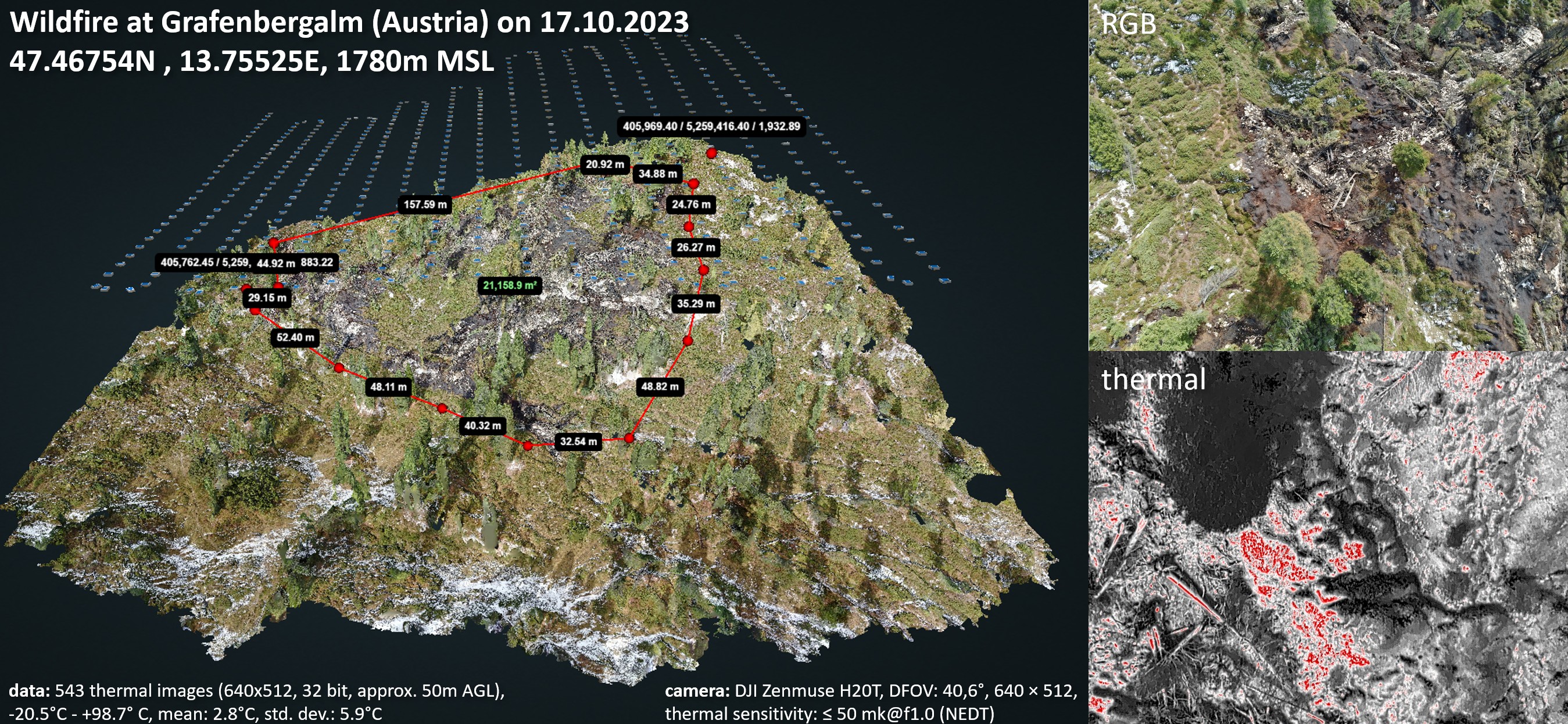}
\caption{A georeferenced 3D reconstruction of the Grafenberg Alm wildfire site, alongside a sample of synchronized RGB and thermal drone images. The 543 individual drone capture poses are marked. Temperature statistics of the dataset and specifications of the thermal sensor are also provided.}\label{Fig:Grafenbergalm}
\end{figure}

A dataset of 543 thermal images was captured by drone during a wildfire at Grafenberg Alm, Austria (1,780 m above sea level, ASL) on 17th of October 2023 at approximately 2pm local time. The ambient temperature was approximately 9$^\circ C$ (cf. Fig. \ref{Fig:Grafenbergalm}). The fire's location near the tree line resulted primarily in observed surface and ground fires with minimal occluding vegetation.

This dataset, however, reflects only the specific ambient conditions present during its recording. To increase its size and variability for more robust model generalization, we embed a latent diffusion model (LDM) \cite{rombach2022high} into a vector quantized variational autoencoder (VQ-VAE) \cite{van2017neural}
(See Supplementary Fig. 1 for additional details on the generating surface temperature model architecture).

\begin{figure}[H]
\centering
\includegraphics[width=1.0\textwidth]{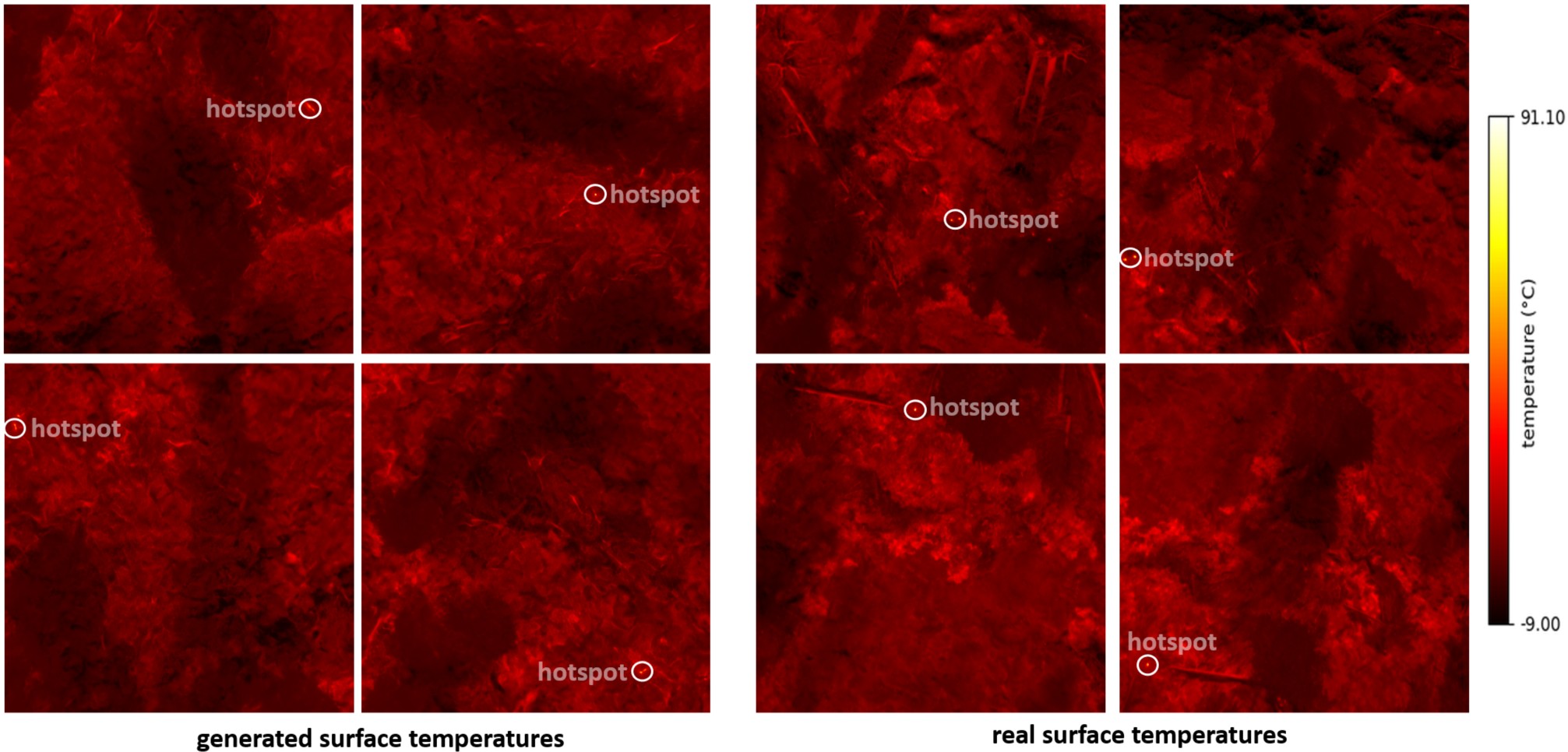}
\caption{A comparison between synthetic surface temperature generated by the proposed VQ-VAE/LDM model (left) and real thermal drone recordings from the Grafenberg Alm wildfire (right). Temperatures are color-coded according to the same scale. The generated samples reproduce similar visual structures and temperature distributions of ground features and fire hotspots observed in real measurements, demonstrating the realism of the synthesized data.}\label{Fig:gen_vs_real}
\end{figure}

This model was used to generate a total of 75,000 new thermal images of unconcluded surface temperatures at 9$^\circ C$ ambient temperature (the same ambient temperature at which the real thermal images have been recorded). The VQ-VAE encoded images by compressing them from 512×512 pixels to a 64×64 latent representation and then reconstructed them back to the original resolution. This compression enabled efficient training under limited computational resources while maintaining image fidelity. The diffusion process was applied in the latent space via the LDM rather than directly in pixel space, balancing computational efficiency with generative quality. 
Supplementary Fig. 2 presents the training analysis with the evolution of the loss functions, demonstrating stable convergence for both VQ-VAE and LDM. In Figure  \ref{Fig:gen_vs_real} , we present a qualitative comparison between two independent datasets: generated samples and real observations. The images do not represent one-to-one correspondences, as they originate from different datasets. Instead, the comparison illustrates that the generated outputs reproduce similar structural characteristics and fine-scale patterns observed in real data, demonstrating the model’s ability to capture the underlying data distribution and produce realistic samples.

The generated images feature locally varying surface temperatures and fire conditions but are limited to the original dataset's ambient temperature of 9$^\circ C$ and maximum fire temperature of 98$^\circ C$. To extend this range to other ambient temperatures and higher fire temperatures, we apply the following temperature augmentation model:

Field experiments \cite{kreye2018effects} have shown that biomass on the forest ground can, due to solar heating, reach temperatures of no more than 15$^\circ C$ above ambient. This process is called direct sunlight absorption. Any temperature above this level can indicate a potential fire, although firefighters use a higher thresholds (usually 50-60$^\circ C$) to confirm active surface or ground fires. 

We conservatively classified surface points with temperature $t$ into two categories: non-fire (if $T_{\text{lower}} < t \leq T_{\text{upper}}$) and potential fire (if $t > T_{\text{upper}}$), where $T_{\text{upper}}$ is defined as the ambient temperature plus $15^\circ C$ and $T_{\text{lower}}$ is the freezing point of $0^\circ C$. For both our original and VQ-VAE/LDM-augmented data, the ambient temperature was constant at $9^\circ$C, resulting in $T_{\text{upper}} = 24^\circ C$.

For each none-fire point, an offset temperature $\Delta T$ is applied that is equivalent to the desired new ambient temperature minus the reference ambient temperature of our original data ($9^\circ C$). However, a naive implementation of this augmentation scheme resulted in discontinuities at the boundaries of $T_{lower}$ and $T_{upper}$. To avoid  this, we apply a smooth weighting based on the sigmoid function $\psi(t)$:

 \begin{equation}
  \label{eq:non_fire_augmentation}
  \begin{split}
  t_{new} & = t + w(t)\cdot \Delta T,\\
  w(t) & = \psi\left(\alpha \cdot (t - T_{lower})\right) - \psi\left(\alpha \cdot (t - T_{upper})\right), 
  \end{split}
  \end{equation}
 
where $\alpha$ controls the steepness of the transition between valid and excluded temperature ranges. By adjusting $\alpha$, the smoothness of the augmentation boundary can be tuned. Empirically, we chose $\alpha$=0.5. 
Supplementary Fig. 3 illustrates the effect of different $\alpha$ values on the transition behavior. 

Each fire point is scaled as follows:

 \begin{equation}
  \label{eq:fire_augmentation}
  \begin{split}
  t_{new} & = T_{upper} + s \cdot (t - T_{upper}),\\
    s &= \frac{T'_{max} - T_{upper}}{T_{max} - T_{upper}},
  \end{split}
  \end{equation}

where $T_{max}$ is the highest fire temperature in the original data ($98^\circ C$), and $T'_{max}$ is the target maximum fire temperature for augmentation (we chose $T'_{max}=300^\circ C$). 

Supplementary Fig. 4 illustrates the temperature augmentation scheme, transforming a sample from our original $9^\circ C$ ambient and a maximum fire temperature of $53^\circ C$ in this sample to the new ranges of $0^\circ C$ to $30^\circ C$ for ambient temperature and $50^\circ C$ to $300^\circ C$ for fire temperature (in four discrete steps each).

Scaling of the fire temperatures is pre-computed by applying Eqn. \ref{eq:fire_augmentation} to the 75,000 surface temperate images that are generated by our VQ-VAE/LDM model. However, instead of pre-computing augmented non-fire temperatures, we integrate this augmentation (Eqn. \ref{eq:non_fire_augmentation}) directly into the procedural forest simulation (Sect. \ref{Sec:ProceduralForest}), computing and combining surface and vegetation temperatures simultaneously for each ambient temperature in 1°C increments.

\subsection{Adding Procedural Forest}\label{Sec:ProceduralForest}
We add simulated forest environments on top of generated surface temperatures using a procedural generation algorithm integrated into a custom GAZEBO-based drone simulator (see Sect. \ref{Sec:Methods} for details). This simulator supports far-infrared imaging with thermal cameras. The procedurally generated trees are parameterized to approximate a European forests (see Sect. \ref{Sec:Methods} for details).

For our training process, learning global occlusion behavior is essential rather than understanding local occlusion patterns. Specifically, the objective is to learn the broader radiometric characteristics associated with vegetation-induced occlusion rather than memorizing specific local canopy structures or scene-dependent features. Therefore, climatic and geographic environmental factors, including species composition, foliage characteristics, and seasonal conditions, were intentionally held constant during the simulation process.



We varied four key simulation parameters to generate training and test data : forest density, ambient  temperature, solar azimuth, and the additional temperature increase from direct sunlight absorption \cite{kreye2018effects} (see Supplementary Table 1 for additional details on the simulation parameters). For each simulated drone flight, these parameters were selected uniformly at random from predefined ranges and discrete intervals.

For a chosen ambient temperature, we randomly select a surface temperature image from the data pool of 75,000 sample (generated for 9$^\circ C$ ambient temperature using our VQ-VAE/LDM model) and augment it to the chosen ambient temperature, as explained Section \ref{Sec:GeneratingSurfaceTemperatures}). The resulting surface temperature image is then mapped onto the simulated forest ground. We then populate it with a procedural forest of the selected density. The thermal radiation is simulated based on the ambient temperature, solar direction, and direct sunlight contribution. Each simulated drone flight generates a set of thermal images captured at the waypoints defined by the synthetic aperture (Sect. \ref{Sec:AOS}). The corresponding dataset sizes (number of simulated images) for our training and test sets are also included for both one-dimensional and two-dimensional synthetic apertures (see Sect. \ref{Sec:AOS} for details). For the 1D SA configuration (11 × 1), the training dataset contains 55,000 images (5,000 samples), and the test dataset contains 26,400 images (2,400 samples). For the 2D SA configuration (11 × 11), the training dataset comprises 600,000 images (5,000 samples), and the test dataset comprises 288,000 images (2,400 samples). 

Note that the training and test datasets are not equally large and use slightly different parameter ranges to ensure computationally feasible simulation times across the entire four-dimensional parameter space. Specifically, the forest density was held constant for the training dataset. We assume our model will learn to generalize from the locally varying occlusion densities present around individual trees. This assumption will be validated in Sect. \ref{Sec:ReconstructingSurfaceTemperatures} by analyzing the reconstruction results from the test dataset, for which the forest density was varied globally.

\begin{figure}[H]
\centering
\includegraphics[width=1.0\textwidth]{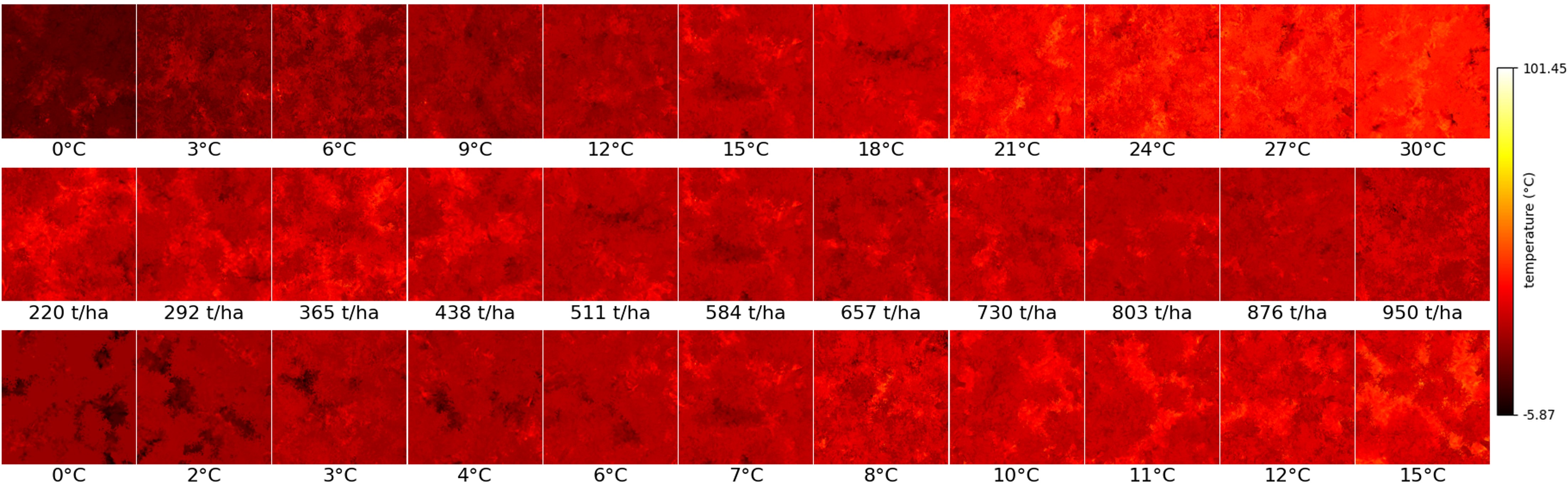}
\caption{Examples for simulated aerial images showing procedural forest on top of generated surface temperatures: Forest density and direct sunlight absorption are fixed to 584t/ha and 7$^\circ C$ while ambient temperature is changed (top). Ambient temperature and direct sunlight absorption are fixed to 15$^\circ C$ and 7$^\circ C$ while forest density is changed (center). Forest density and ambient temperatures are fixed to 584t/ha and 15$^\circ C$ while direct sunlight absorption is changed (bottom). Temperatures are color-coded according to the scale. Solar azimuth was nadir (0$^\circ$) in all cases.}\label{Fig:simulation_results}
\end{figure}

Figure \ref{Fig:simulation_results} illustrates simulated aerial image examples of procedural forests on top of generated surface temperatures for different vegetation densities, ambient temperatures, and solar azimuths.

Supplementary Fig. 5 illustrates visual comparisons between simulated and real thermal aerial images of wildfire below forest canopy.
\subsection{Suppressing Occlusion with Airborne Optical Sectioning}\label{Sec:AOS}
The Airborne Optical Sectioning (AOS) imaging method \cite{schedl2020airborne, schedl2020search, schedl2021autonomous, kurmi2018airborne, bimber2019synthetic, kurmi2019statistical, kurmi2019thermal, kurmi2020fast, kurmi2021pose, kurmi2106combined, nathan2022through, youssef2025deepforest, nathan2024autonomous, nathan2024reciprocal, youssef2024fusion, kerschner2024stereoscopic, amala2023synthetic, amala2023drone, seits2022evaluation, amala2022inverse} applies a special signal processing principle known as synthetic aperture (SA) sensing. Today, synthetic aperture (SA) sensing is used in many fields, such as radar \cite{moreira2013tutorial, li2015synthetic, rosen2002synthetic}, interferometric microscopy \cite{ralston2007interferometric}, sonar \cite{hayes2009synthetic}, ultrasound \cite{jensen2006synthetic, zhang2016synthetic}, LiDAR \cite{barber2014synthetic, turbide2017synthetic}, imaging \cite{yang2016kinect, pei2019occluded}, and radio telescopes \cite{levanda2009synthetic, dravins2015optical}. Similar to interlinking distributed radio telescopes to improve measurement signals by coherently combining individual receptions, AOS integrates optical images recorded over a large forested area in order to computationally remove in real time occlusion caused by trees. This creates extremely shallow depth-of-field integral images with a largely unobstructed view of the forest floor. 

\begin{figure}[H]
\centering
\includegraphics[width=0.8\textwidth]{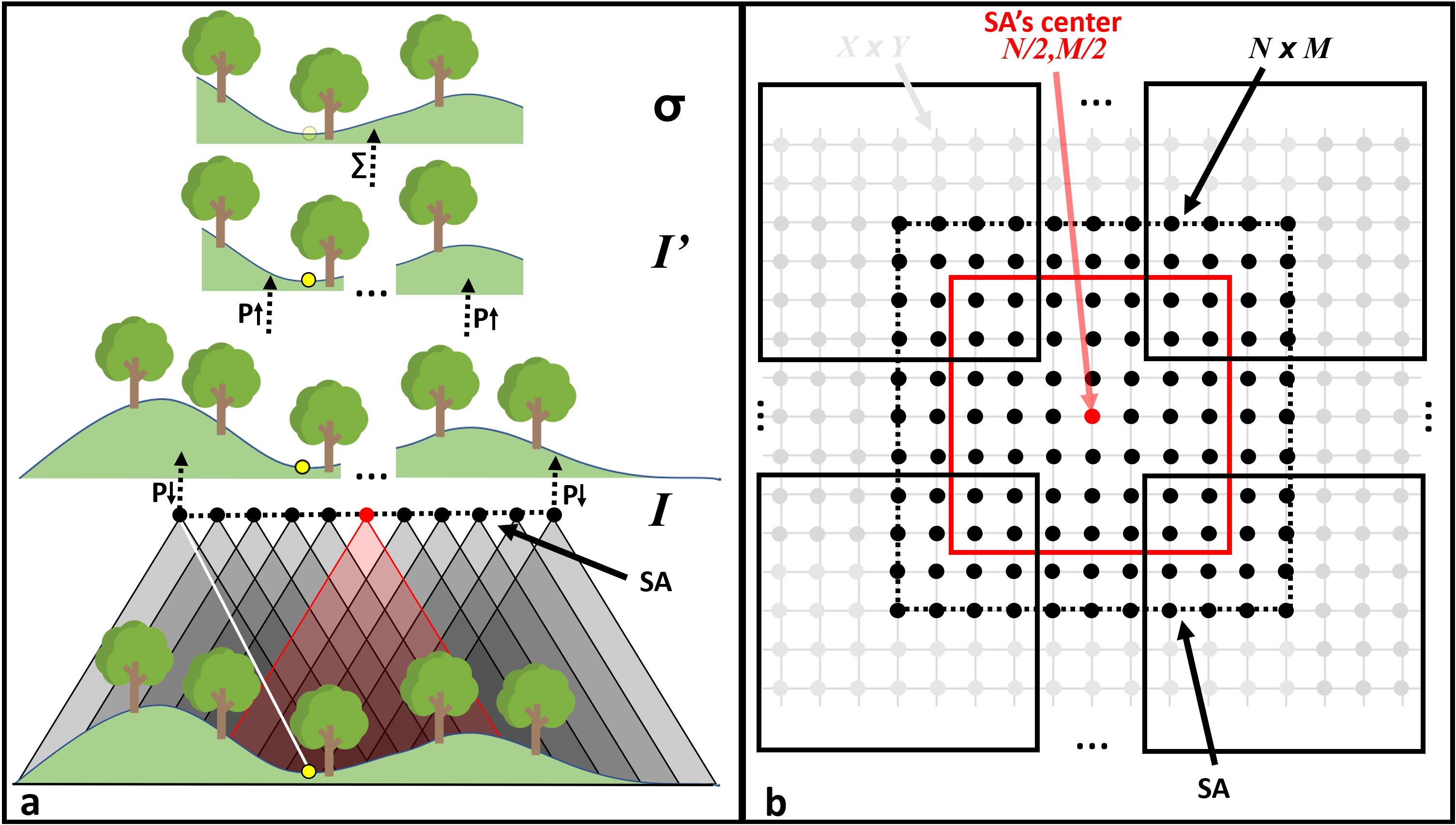}
\caption{The AOS principle: \textbf{a}, Images ($I$) captured at waypoints (black dots) along a synthetic aperture (SA) are projected ($P\big\downarrow$) onto a digital elevation model of the ground surface, back-projected ($P\big\uparrow, I'$) to the SA's central perspective (red dot), and integrated ($\Sigma$). The signal from a heavily occluded surface point (yellow dot) appears in the resulting integral image ($\sigma$) if it is visible from at least one sampling waypoint—even if only weakly, depending on the degree of occlusion. \textbf{b}, To cover an arbitrarily large area, the SA's sampling grid (black and red dots) can be shifted across an extended grid (gray dots). Each shift produces an integral image from a new perspective. The black squares illustrate the minimal required image overlap between the SA's corner waypoints and the integral image at the SA's center (red square).}\label{Fig:AOS_principle}
\end{figure}

In this work, occlusion removal using AOS is implemented as follows (cf. Fig. \ref{Fig:AOS_principle}):

A grid of $N \times M$ drone waypoints ($n,m$) is defined, representing the synthetic aperture (SA). Thermal images ($I_{n,m}$) are captured at each waypoint. The processing pipeline for these images is as follows (cf. Fig. \ref{Fig:AOS_principle}\textbf{a}):

\begin{equation}
I'_{n,m}=P\big\uparrow_{n/2,m/2}\left(P\big\downarrow_{n,m}\left(I_{n,m}\right)\right), \label{Eqn:ImageTransformation}
\end{equation}

where $P\big\downarrow_{n,m}$ denotes the perspective projection of an image from waypoint $(n,m)$ onto a digital elevation model (DEM) of the ground surface, and $P\big\uparrow_{n/2,m/2}$ is the subsequent back-projection from the DEM to the central waypoint of the grid $(n/2, m/2)$, representing the SA's center. 

The AOS integration is then performed as a per-pixel operation:

\begin{equation}
\sigma^{N,M}=\frac{1}{NM}\sum\limits_{n=1,m=1}^{N,M}I'_{n,m}, \label{Eqn:AOS}
\end{equation}

where $\sigma^{N,M}$ is the occlusion-suppressed integral image for the SA's center. Note that image regions with non-overlapping back-projections must be handled appropriately. In contrast to zero-padding, which attenuates signal strength during integration, we entirely exclude non-overlapping pixels in Eqn. \ref{Eqn:AOS}. This approach preserves signal strength but, like padding, results in a lower effective sampling density in areas with less overlap.

For applying a fixed SA of size $N\times M$ to an arbitrarily larger grid of $X \times Y$ drone waypoints ($X \geq N, Y \geq M$), Eqn. \ref{Eqn:AOS} can be used in a sliding window manner (potentially with strides $>1$ in both directions), as illustrated in Fig. \ref{Fig:AOS_principle}\textbf{b}:

\begin{equation}
\sigma^{N,M}_{x,y}=\frac{1}{NM}\sum\limits_{i=-N/2,j=-M/2}^{N/2,M/2}I'_{x+ i,y+j}. \label{Eqn:Sliding_AOS}
\end{equation}

This results in a maximum of $X \times Y$ integral images (fewer with strides $>1$), one for each possible waypoint perspective ($x,y$).
Note that the waypoint grid must be padded to facilitate the computation of integral images for waypoints near or at the grid's borders. As stated above, we exclude non-overlapping pixels in Eqn. \ref{Eqn:AOS} to preserve signal strength.

\begin{figure}[H]
\centering
\includegraphics[width=1.0\textwidth]{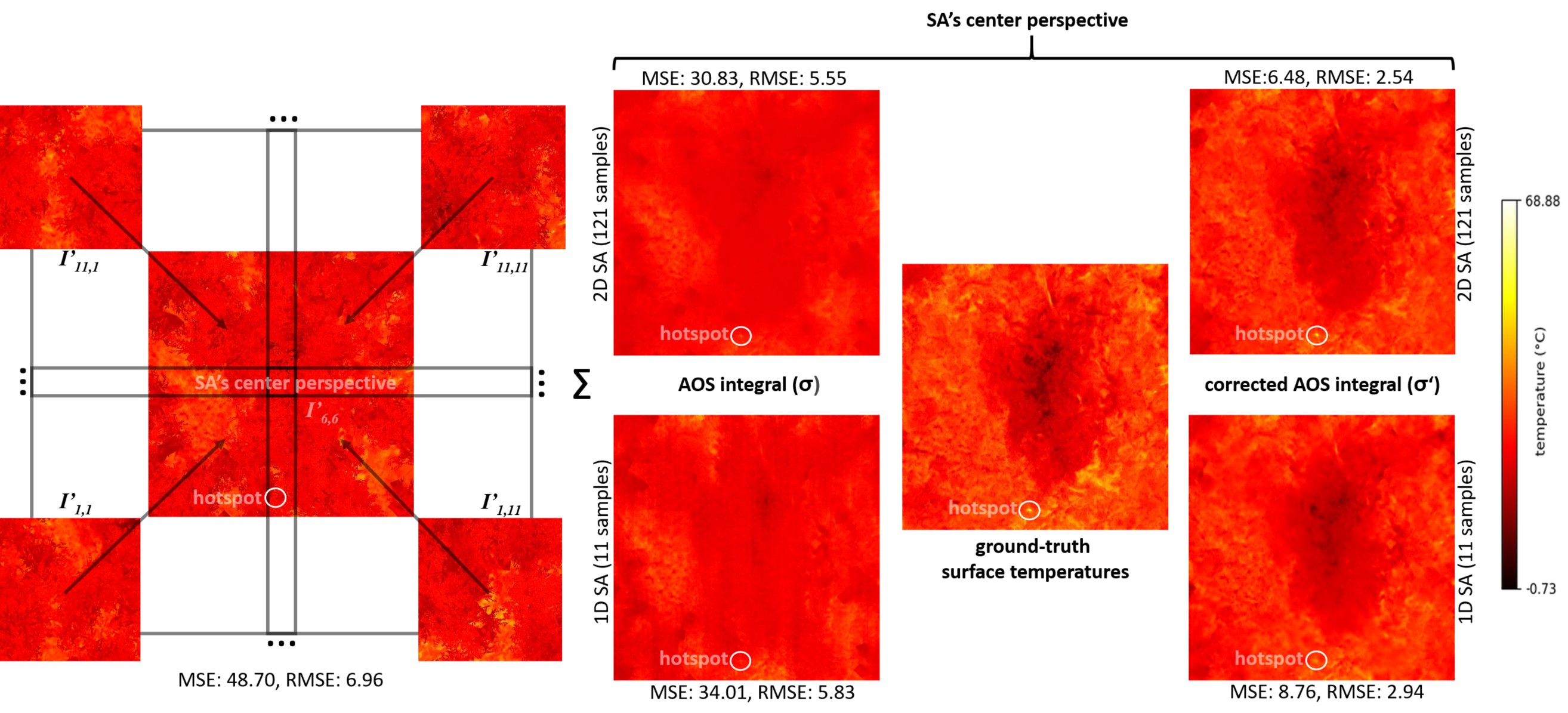}
\caption{AOS example (left to right): Single thermal images (shown here: the center perspective and the overlapping regions of back-projected corner perspectives) from an $11 \times 11$ SA (2m sampling distance, 35m AGL) reveal temperatures of tree crowns and the partially occluded forest surface. The AOS integral ($\sigma$), computed by combining the SA samples (all 121 for the 2D grid; only the vertical 11, $I'_{6,1}..I'_{6,11}$, for the 1D strip), partially suppresses the occlusion. A comparison with the ground-truth surface temperatures (without forest vegetation) highlights the limitations of AOS under dense occlusion. Our new approach (Sect.\ref{Sec:ReconstructingSurfaceTemperatures}) corrects the AOS integral ($\sigma'$), enabling a much more accurate recovery of surface temperatures. All results are simulated for a $15^\circ C$ environment and a forest density of 700 trees/ha (Sects. \ref{Sec:GeneratingSurfaceTemperatures} and \ref{Sec:ProceduralForest}). Temperatures are color-coded according to the scale. A ground fire hotspot (white circle) is detectable in the corrected AOS integral but only weakly in the uncorrected version. Compared to the ground truth, the mean square error (MSE) and root mean square error (RMSE) decrease from the single-image measurements to the AOS integral and drop further for the corrected AOS integral. Results from the 2D SA are generally superior to those from the 1D SA due to higher and more directionally varied sampling.}\label{Fig:AOS_example}
\end{figure}

Since AOS is most efficient when sampling just above the treeline \cite{kurmi2019statistical}, we selected an SA altitude of 35m above ground level (AGL). This assumes a maximum tree height of 30m and includes a 5m safety margin. From this altitude, our drone's thermal camera covers a $22\mathrm{m} \times 22\mathrm{m}$ area on a planar ground surface. To ensure a minimal ground overlap of 2m between the SA's border perspectives and its center perspective, and to avoid oversampling \cite{kurmi2019statistical}, we chose $N=11$ and $M=11$ with waypoint distances of 2m. Consequently, our SA measures $20\mathrm{m} \times 20\mathrm{m}$, while the ground coverage of the integral images remains $22\mathrm{m} \times 22\mathrm{m}$. Figure \ref{Fig:AOS_example} illustrates an example.

We have also experimented with one-dimensional synthetic apertures (i.e., sampling along waypoint strips instead of grids). The same fundamental principles apply, but Eqns. \ref{Eqn:ImageTransformation} - \ref{Eqn:Sliding_AOS} must be adapted from 2D grids to 1D strips \cite{schedl2021autonomous}. For the 1D SA case, we use $N=11$ with 2m waypoint distances, which still results in a $22\mathrm{m} \times 22\mathrm{m}$ ground coverage area. However, the integral images are only sampled in one dimension (i.e., along a single direction) and with 11 samples instead of 121. Consequently, the results from 2D SAs are generally superior to those from 1D SAs due to higher and more directionally varied sampling (cf. Fig. \ref{Fig:AOS_example}). The primary trade-off is that capturing more samples sequentially (i.e., with a single drone) requires more time.

A fundamental limitation of occlusion removal techniques like AOS is that the unregistered signal from out-of-focus occluders (e.g., forest vegetation) is not eliminated but blurred across the integral image. This suppresses, but does not eliminate, local occlusion. In contrast, the registered signal from in-focus regions (e.g., the forest ground) is amplified. However, in dense vegetation, this blurred occlusion signal remains strong. When integrating thermal images, where signal corresponds to temperature, the desired ground surface temperatures are consequently biased by the superimposed, blurred temperatures of the vegetation. This effect is illustrated in Fig. \ref{Fig:AOS_example}. Nevertheless, the accurate surface temperature signal persists within this mixture and may be recoverable. The following section presents a restoration framework to reconstruct surface temperatures from AOS integral images. Examples of this correction are provided in Fig. \ref{Fig:AOS_example}.

\subsection{Reconstructing Surface Temperatures}\label{Sec:ReconstructingSurfaceTemperatures}

To reconstruct surface temperatures ($\sigma'$) from AOS integral images ($\sigma$), we employ a visual state space model (See Supplementary Fig. 6 for additional details on the surface temperature reconstruction model architecture) -- specifically the VmambaIR framework \cite{10843251}, where the standard Mamba \cite{gu2024mamba} blocks were replaced by MambaOut modules \cite{Yu_2025_CVPR}. During training, the ambient temperature is treated as a known parameter derived from the training data and embedded throughout the network layers to improve reconstruction accuracy. During inference, however, it is estimated directly from the observations by computing the mean thermal radiance across all frames within the SA scanning area. We assume that local environmental effects -- such as canopy shading, topography, and soil moisture variations -- are sufficiently captured by this spatial averaging.

\begin{figure}[H]
\centering
\includegraphics[width=1.0\textwidth]{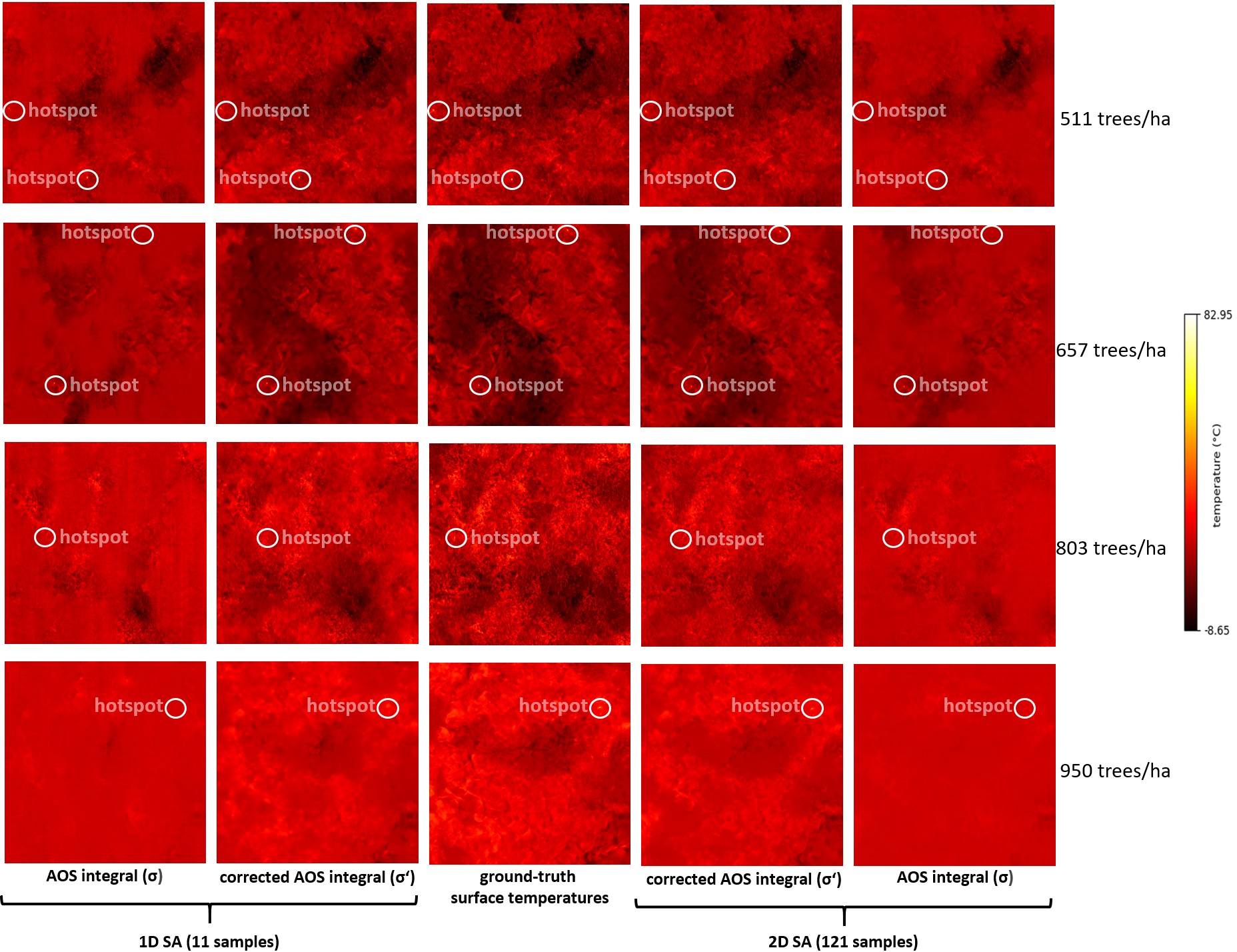}
\caption{Simulated correction results for test data samples of varying forest densities: AOS integral images ($\sigma$) for 1D and 2D SAs, the corresponding corrected AOS integrals ($\sigma'$), and the ground-truth surface temperatures (i.e., without occlusion forest vegetation). Ambient temperature and direct sunlight absorption are fixed to 15$^\circ C$ and 7$^\circ C$ while forest density is changed. Note, that the models were trained exclusively with a fixed forest density of 220t/ha.  Temperatures are color-coded according to the scale.}\label{Fig:local_density}
\end{figure}

We trained two separate versions of this model: one for the $11 \times 1$ 1D SA and another for the $11 \times 11$ 2D SA.  
Supplementary Fig. 7 presents the training analysis with the evolution of the loss functions, demonstrating stable convergence. The training data  was split into 80\% for training and 20\% for validation. As stated earlier, we used a dataset with a fixed forest density of 220t/ha to reduce complexity. Results demonstrated that local density variations around trees (e.g., denser regions near stems versus more open areas in the crown peripheries) were sufficiently distributed in the data to generalize the surface temperature restoration under varying dense forest occlusion. Examples of these local density variations and corrections with our models are illustrated in Figure \ref{Fig:local_density}.

\begin{figure}[H]
\centering
\includegraphics[width=1.0\textwidth]{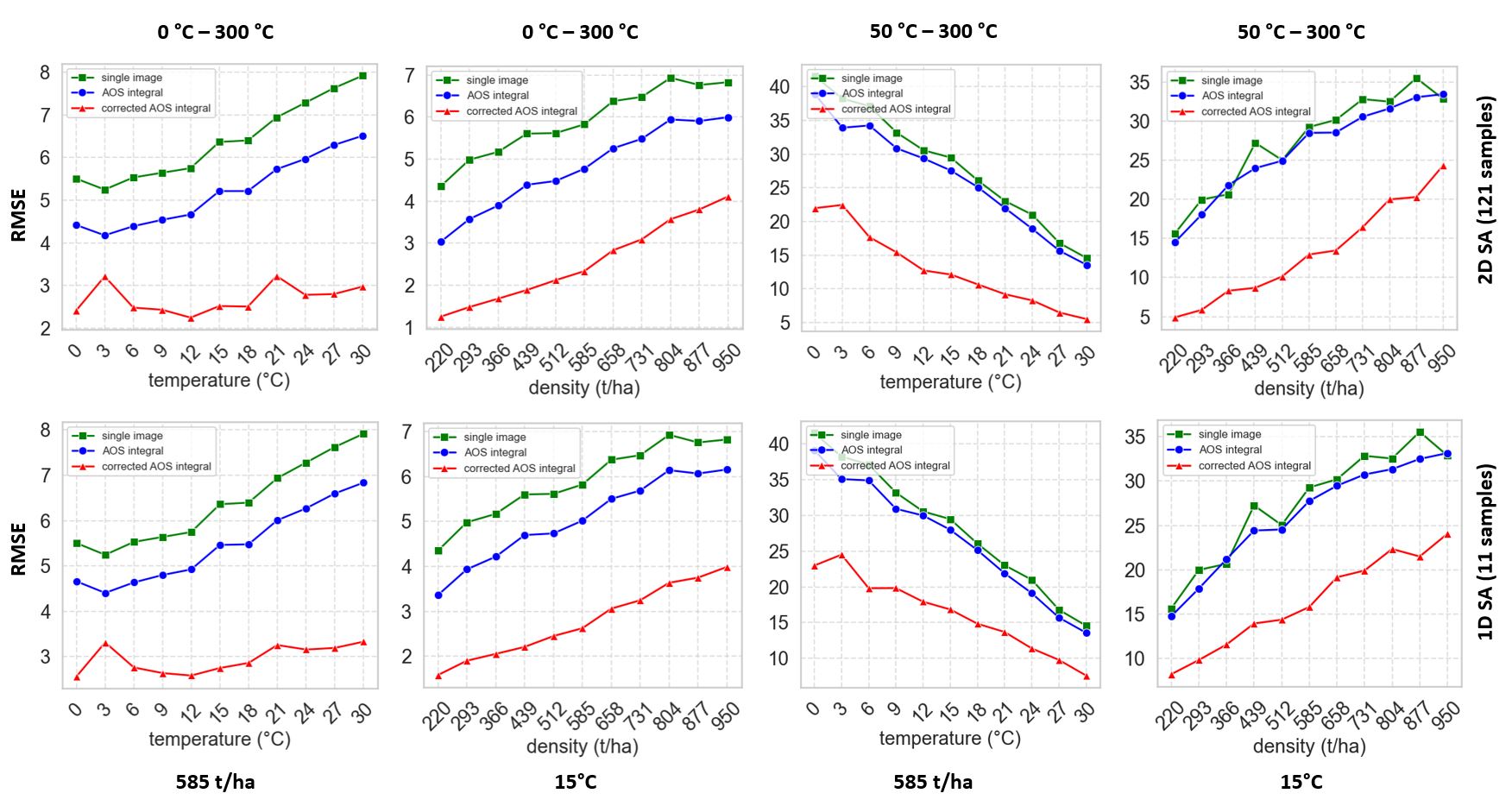}
\caption{Quantitative evaluation on test dataset: Shown are plots of the average Root Mean Squared Error (RMSE) for both 1D and 2D SA cases. The AOS integrals, corrected AOS integrals, and single thermal aerial images (sharing the same center perspective as the integrals) are all compared against the ground-truth surface temperatures (without occluding forest vegetation). For scenarios where ambient temperature was varied, the forest density was fixed at 585t/ha; conversely, when forest density was varied, the ambient temperature was fixed at 15$^\circ$C. All averages include the full range of solar azimuth and direct sunlight absorption variations. While the left two columns consider the full temperature range (0$^\circ C$ - 300$^\circ C$), the right two columns focuses only on the high temperature regime of confirmed active fire (50$^\circ C$ - 300$^\circ C$).}\label{Fig:evaluation_restoration}
\end{figure}

While Figure \ref{Fig:local_density} shows individual visual examples, Figure~\ref{Fig:evaluation_restoration} presents a qualitative evaluation on our full test datasets.


The results in Figure \ref{Fig:evaluation_restoration} lead to the following conclusions: 1.) AOS integral images improve surface temperature estimation over conventional aerial thermal imaging by removing occlusion, but our correction models yield far superior results. 2.) 2D SA integration outperforms 1D SA integration due to its higher sampling rate and directionally varying measurements. 3.) All methods become more error-prone as forest density increases. 4.) Temperatures in higher ranges (e.g., the confirmed active fire regime above 50$^\circ C$) undergo more severe corrections and exhibit a higher RMSE than temperatures in lower ranges. 5.) While the estimation of lower temperatures generally suffers more from increasing ambient temperatures, the estimation of higher temperatures can improve with presence of sufficient occlusion. This may seem paradoxical, but it is due to the residual temperature error from occlusion removal, which shifts the estimates a bit closer to the ground truth.   


 
 

For our simulated data, the corrected AOS integrals significantly reduced the RMSE. For 2D SAs, the improvement was 2.5-fold on average (up to 3.5-fold) over conventional aerial thermal images, and 2-fold on average (up to 2.4-fold) over uncorrected AOS integrals. The improvement was slightly lower for 1D SAs, with average gains of 2.2-fold (max 2.8-fold) and 1.9-fold (max 2.4-fold), respectively.

\subsection{Field Experiments}\label{Sec:FieldExperiments}
\begin{figure}[H]
\centering
\includegraphics[width=1.0\textwidth]{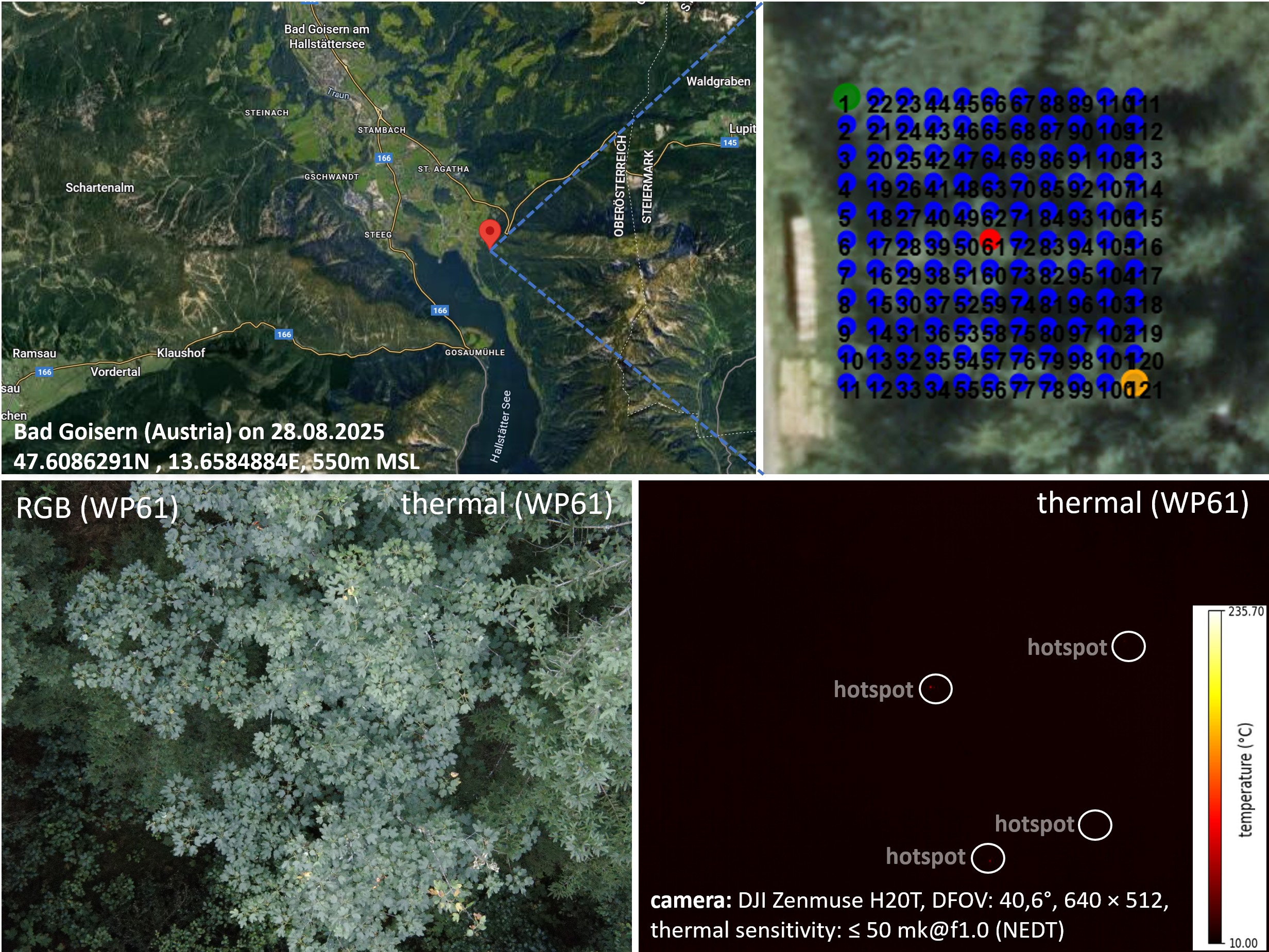}
\caption{Field experiment site near Bad Goisern am Hallstättersee, with a close-up of the 121-waypoint 2D scanning array grid. The center perspective (waypoint 61) is shown using both RGB and thermal aerial imagery with locations of hotspots on the ground. The known locations of four hotspots are shown in the thermal image.}\label{Fig:field_experiment_site}
\end{figure}

In addition to the simulated results in Section \ref{Sec:ReconstructingSurfaceTemperatures}, we present results from real thermal drone images of occluded surface fires captured during field experiments. These experiments were carried out in collaboration with firefighters and took place on the 28th of August 2025 (first drone flight: 7:39PM-7:50PM, second drone flight: 8:00PM-8:11PM) in a mixed forest near Bad Goisern am Hallstättersee in Austria (cf. Fig. \ref{Fig:field_experiment_site}).  

Surface fire was simulated using glowing charcoal in four metal bowls placed randomly on the ground. The forest was scanned by the drone with a 2D SA of the same properties as those used for training our models ($11 \times 11$, 2m sampling distance, 35m AGL). The experiment was repeated twice under the same scanning conditions but with different fire temperatures and ground positions.

In both cases, we estimated an ambient temperature of $13^\circ C$ by averaging the temperatures of all pixels from all 121 thermal images. The ground-truth temperatures of the hotspots were obtained by recording them unoccluded with the drone's thermal camera and then averaging the corresponding pixel temperatures. Note, that we assume a radiometrically calibrated thermal camera.

\begin{figure}[H]
\centering
\includegraphics[width=0.9\textwidth]{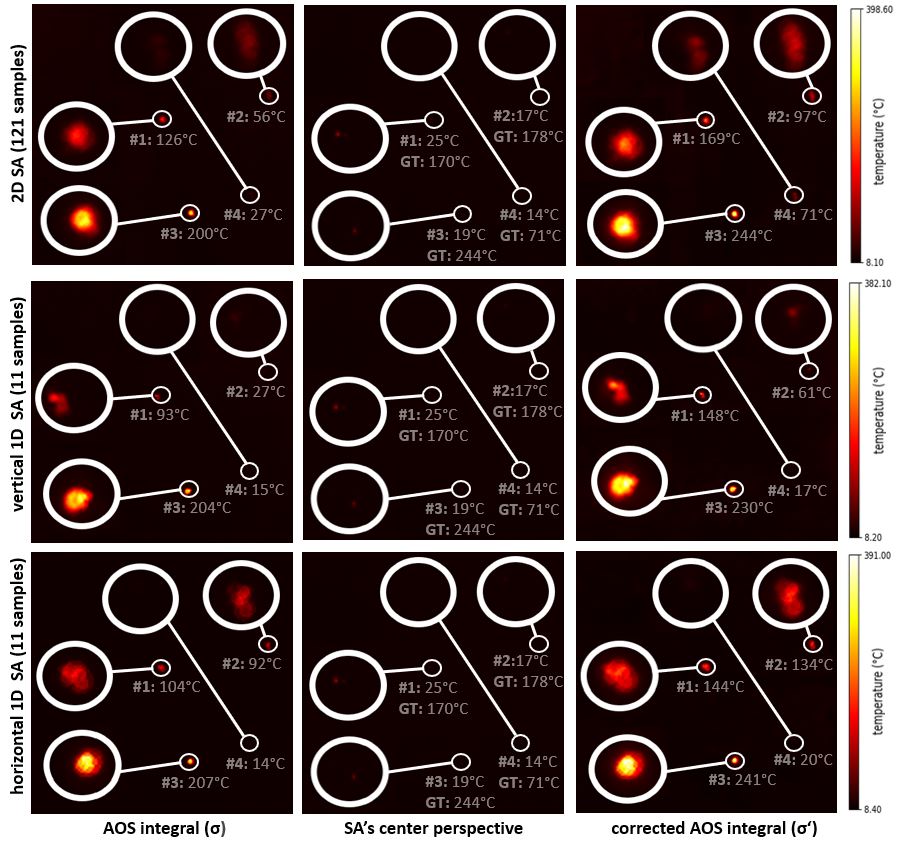}
\caption{Field experiment 1 involved four simulated surface fires created using glowing charcoal in metal bowls placed beneath forest vegetation (close-ups are shown). The results present three types of thermal imagery: regular thermal images from the SA's center perspective waypoint (center column) showing average hotspot temperatures (both ground truth (GT) and image-derived); AOS integral images (left column) with average hotspot temperatures determined from the integral images; and corrected AOS images (right column) with average hotspot temperatures obtained from the corrected integral images. Sampling was performed using three approaches: 2D SA (top row), 1D SA as a vertical strip of the 2D SA (center row), and 1D SA as a horizontal strip of the 2D SA (bottom row). Temperatures are color-coded according to the scales. Ambient temperature was $13^\circ C$.}\label{Fig:field_experiment_1}
\end{figure}

Figure \ref{Fig:field_experiment_1} presents the results of the first experiment. Assuming a common fire detection threshold of $50-60^\circ C$, none of the hotspots would have been detectable in a regular thermal image captured from the SA's center perspective due to occlusion. Although fragments of hotspots are partially unoccluded from other perspectives, they never reveal the complete hotspot footprints. These fragments are combined in the AOS integral images, but the resulting temperatures are too low because cooler temperatures from occluding vegetation are also averaged. Based on the assumed fire threshold, 2-3 hotspots could have been clearly detected in AOS integral images. However, with the corrected AOS integral sampled by the 2D SA, all four hotspots could have been detected. The corrected temperatures are closer to the ground truth values compared to the uncorrected temperatures. In general, 2D SA sampling outperforms 1D SA sampling, and the corrected integrals perform better than the uncorrected ones. Differences in the strip direction (horizontal or vertical) of the 1D SAs are attributed to local occlusion conditions, meaning that hotspots might be more visible from one direction than the other.  

\begin{figure}[H]
\centering
\includegraphics[width=0.9\textwidth]{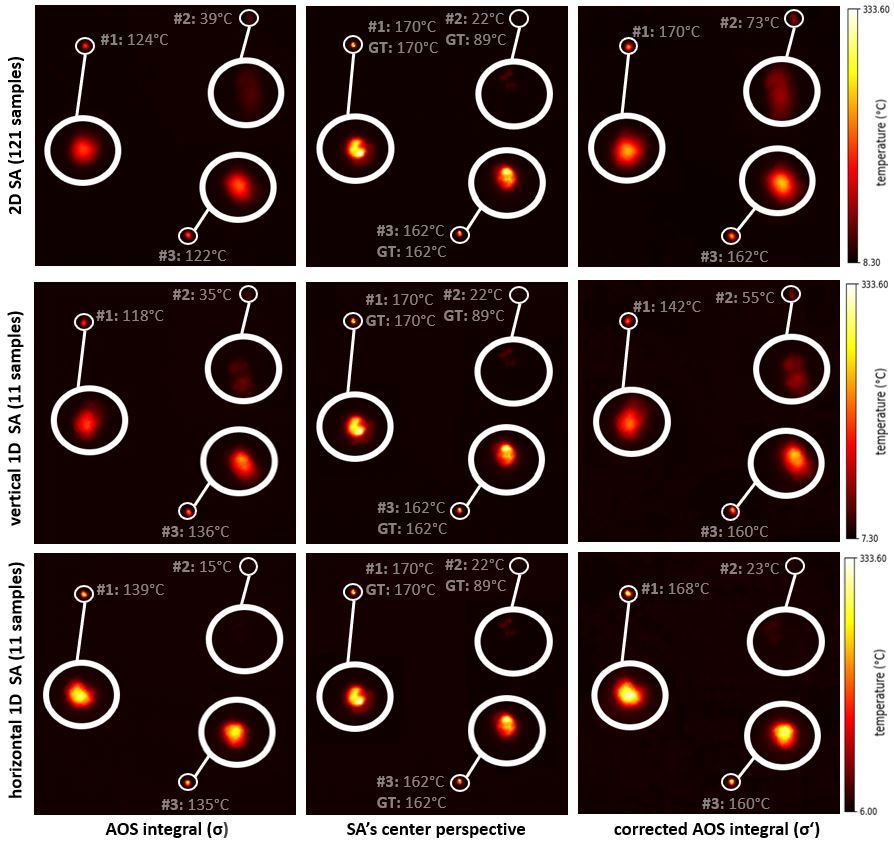}
\caption{For field experiment 2, simulated surface fires were reposition and cooled down (close-ups are shown). The results present three types of thermal imagery: regular thermal images from the SA's center perspective waypoint (center column) showing average hotspot temperatures (both ground truth (GT) and image-derived); AOS integral images (left column) with average hotspot temperatures determined from the integral images; and corrected AOS images (right column) with average hotspot temperatures obtained from the corrected integral images. Sampling was performed using three approaches: 2D SA (top row), 1D SA as a vertical strip of the 2D SA (center row), and 1D SA as a horizontal strip of the 2D SA (bottom row). Temperatures are color-coded according to the scales. Ambient temperature was $13^\circ C$.}\label{Fig:field_experiment_2}
\end{figure}

In the second experiment (Fig. \ref{Fig:field_experiment_2}), three of the four metal bowls were randomly repositioned to create different occlusion conditions. One bowl, having cooled to ambient temperature, was removed. The charcoal in the three remaining bowls had also cooled significantly since the first experiment. Based on our fire detection threshold, two hotspots were detectable in all cases (regular thermal images, AOS integrals, and corrected AOS integrals) because they were largely unobstructed. The third, more heavily occluded hotspot, however, remained below the fire threshold in all cases except for the corrected AOS integrals, where its measured value was above the threshold. Consistent with the results of the first experiment, 2D SA sampling demonstrates superior performance to 1D SA sampling. 

For the first and second flights using 2D SA sampling, the average RMSE gains for corrected AOS integrals over conventional thermal images were 7.6-fold and 1.8-fold, respectively (with maximums of 12.8 and 3.3). When comparing corrected to uncorrected AOS integrals, the average gains were 1.7-fold and 1.5-fold (maximums of 2.6 and 1.9). For 1D SA sampling, the gains for corrected AOS integrals over conventional images were 6.1-fold and 1.2-fold on average (maximums of 12.7 and 2.5), while the gains for corrected over uncorrected integrals were 1.5-fold and 1.3-fold (maximums of 2.3 and 1.6).

 
 
 
 


Supplementary Fig. 8 presents results of a third field experiment, demonstrating our approach's performance in a simulated search-and-rescue scenario with people rather than hotspots on the forest ground. In this application, the heat signatures of people were significantly cooler than those of surface fires.

\section{Discussion}\label{Sec:Discussion}
We have presented an approach for estimating ground surface temperatures from aerial thermal imagery through occluding forest foliage. This is achieved by combining synthetic aperture sensing with deep learning. It enables a range of critical applications, most notably the early detection of ground and surface fires. Beyond fire detection, the method is potentially also applicable to other domains, such as search and rescue \cite{schedl2020search, schedl2021autonomous} and soil water content estimation \cite{casamitjana2020soil, gao2022inversion, ge2021estimating, bertalan2022uav, seo2020soil, zhang2023evaluating}. Long-endurance autonomous drones equipped with this technology could continuously monitor vast forested areas and trigger early alerts upon detecting thermal anomalies. While the results presented in this work used a limited-sized SA, the method itself is scalable to arbitrary sizes by employing a sliding window imaging approach. This applies for both 1D and 2D SA sampling. While 1D sampling offers shorter flight times, 2D sampling delivers enhanced results by capturing a more complete set of measurements from all directions.

A simple temperature threshold is often insufficient for effective detection and classification, as it is easily invalidated by sunlight reflection and thermal radiation from the tree canopy and soil. Conversely, the thermal signatures of critical targets, such as incipient ground fires or people, can be very subtle. The shape and extent of heated areas (i.e., their morphological properties) are equally important for accurate classification. Our approach addresses this challenge by using synthetic aperture sensing to suppress occlusion (thereby mitigating canopy-level radiation and reflection) and deep-learning enabled surface temperature reconstruction to recover such subtle thermal signals obscured by residual thermal blur. While the efficiency of our method decreases with higher occlusion density and ambient temperature, it significantly outperforms conventional imaging techniques. On simulated data encompassing a wide range of ambient temperatures, forest densities, and sunlight conditions, our method achieved a 2 to 2.5-fold reduction in RMSE compared to conventional thermal images and uncorrected AOS integral images. In field experiments that explicitly address the complexities of real-world thermal environments, including variations in species composition and the subtle thermal interactions between the canopy and the understory. The improvement was even more substantial, with a 12.8-fold RMSE gain over conventional thermal images and a 2.6-fold gain over uncorrected AOS integrals. Furthermore, the model exhibits moderate sensitivity to errors in ambient temperature estimation. However, due to training across a wide range of ambient temperatures, the model remains robust to small deviations. In practice, errors within a realistic range primarily lead to slight biases in reconstructed temperature values rather than structural degradation of the recovered thermal patterns. Our experiments successfully revealed the complete shape and extent of fire hotspots and people, unlike conventional imaging which fails under occlusion. Performing such validation requires precise ground-truth measurements. However, acquiring such data in real-world conditions is both challenging and risky, as it requires controlled fire setups and the involvement of fire safety professionals to secure surrounding areas and prevent unintended fire spread. In active, unmanaged environments with dense vegetation, obtaining high-resolution ground temperature measurements is infeasible due to canopy occlusion. Consequently, controlled experiments using charcoal bowls are necessary to generate ground-truth data for quantitative evaluation (e.g., RMSE) and to verify morphological reconstruction against known geometric structures. While these experiments provide essential benchmarks, they cannot fully replicate the stochastic dynamics and environmental complexity of real wildfires. Future work will therefore focus on validation across diverse biomes and during active fire-suppression scenarios, where evaluation will likely emphasize qualitative fire detection rather than quantitative temperature reconstruction, as pixel-accurate ground truth is not obtainable under real wildfire conditions.
Furthermore, the proposed approach operates near real time. On our hardware, the integration of 121 images and the subsequent temperature correction (i.e., image restoration inference) takes approximately 500ms. 

A key challenge was training our model with limited real-world data. This data scarcity is not only a practical challenge but also a fundamental physical limitation. In real forest environments, pixel-accurate ground-truth surface temperatures beneath a forest canopy cannot be obtained without physically removing the occluding vegetation. Unlike traditional wildfire detection tasks, which primarily focus on classification (e.g., fire vs no-fire) and can therefore rely on manually annotated real imagery, the objective of the proposed framework involves radiometric reconstruction, requiring supervised learning against occlusion-free surface temperature maps. Consequently, paired observations consisting of canopy-occluded measurements and corresponding occlusion-free ground temperatures are required for training. Furthermore, directly replacing procedurally generated forest environments with real forest recordings is not feasible, since no physically meaningful relationship would exist between real canopy observations and  augmented ground-temperature maps. We overcame this by employing a combination of surface temperature diffusion and vector quantized variational autoencoder, temperature augmentation, and procedural thermal forest simulation. While our approach focuses on far-infrared (thermal) radiation, it can be extended to other wavelengths, such as red-edge or near-infrared, which are critical for agricultural \cite{chen2022real, velez2023beyond} and forest ecology \cite{youssef2025deepforest, velez2023beyond, trouve2022combining, villacres2022construction, camarretta2020monitoring, reinisch2020combining} applications. Although our models are trained primarily on computed data with a limited range of variability and distribution, they demonstrate strong generalization to real-world conditions represented in our field experiments. Nevertheless, additional validation across diverse biomes, vegetation types, and climatic regions will therefore be necessary to establish the broader generalizability of the proposed approach. However, the results of the field experiments show that coverage of all environmental variations may not be required to learn the relationship between vegetation-induced thermal occlusion and aerial measurements of surface temperature. We expect this to hold for other wavelengths as well, providing an important area for future work.


Future work will further improve the generalizability of the proposed approach by incorporating a larger collection of real wildfire ground observations acquired under diverse environmental conditions. These data will serve as an expanded initial dataset for defining the temperature distributions and morphological characteristics of fire patterns. In addition, the current procedural forest simulator can be extended or replaced with more physically realistic ecological simulation frameworks to better represent the complexity of natural environments and their associated thermal interactions.


\section{Methods}\label{Sec:Methods}

\subsection{Drone Recordings}
A DJI Matrice 300 drone, equipped with a DJI Zenmuse H20T thermal camera, was used to record thermal imagery of real fires and people (see Supplementary Data). This includes the surface fire at Grafenberg Alm (used for data generation, Sect. \ref{Sec:GeneratingSurfaceTemperatures}) and the simulated fires and people at Bad Goisern am Hallstättersee (used for field experiments, Sect. \ref{Sec:FieldExperiments}). The camera has a $640 \times 512$px resolution (cropped to $512 \times 512$px), a 32 bit temperature resolution, a 40.6$^{\circ}$ diagonal field of view, and a thermal sensitivity of $\le 50$ mK at f/1.0. The drone was flown automatically via predefined waypoints using DJI's Pilot software, as detailed in Sects. \ref{Sec:GeneratingSurfaceTemperatures} and \ref{Sec:FieldExperiments}.

\subsection{Procedural Forest Simulation}
Synthetic thermal aerial imagery was generated using a GAZEBO-based drone simulator (\url{https://github.com/bostelma/gazebo_sim}, see Supplementary Code). It supports the simulation of thermal radiation and imaging in 3D virtual environments, as well as thermal texture mapping which was used for integrating surface temperatures. The virtual environments consisted of procedurally generated European broadleaf forests, with tree models parameterized to reflect realistic variations. This included tree heights of 5–20m, trunk lengths of 4–8m, trunk diameters of 20–50cm, and leaf sizes of 5–20cm. A seeded random generator controlled tree distribution and similarity to ensure variety while maintaining controlled patterns. All images were captured with a nadir-oriented virtual camera, producing thermal images at $512 \times 512$px resolution. Forest parameters were varied as summarized in Supplementary Table 1. Camera and imaging parameters were set as described in Sect. \ref{Sec:AOS}.

\subsection{Synthetic Aperture Image Integration}
For synthetic aperture image integration, as explained in Sect. \ref{Sec:AOS}, the Airborne Optical Sectioning (AOS) software (\url{https://github.com/JKU-ICG/AOS/}) was used. Simulated and real thermal images, along with their corresponding pose data, were loaded into the software, and the synthetic focal plane was adjusted to the forest ground. For the simulated images, the precise poses were defined and already known. For the real drone images, precise poses were determined using the structure-from-motion and multi-view stereo pipeline COLMAP (\url{https://colmap.github.io/}). While poses from the drone's onboard sensors could have been used, they were less precise.   

\subsection{Training of Models and Inference}
All training and inference processes (see Supplementary Code and Data) were conducted on a system equipped with an Intel Core i9 processor, 64 GB of RAM, and a single NVIDIA RTX 5090 GPU. The AdamW optimizer \cite{loshchilov2017decoupled} was employed with hyperparameters $\beta_{1} = 0.9$ and $\beta_{2} = 0.99$. The learning rate was linearly warmed up to $1 \times 10^{-4}$ over the first 100 iterations and subsequently decayed to $1 \times 10^{-5}$ using a cosine annealing schedule. This configuration was applied for both the generation and reconstruction of surface temperature models, with a batch size of 8. The training objective for the VQ-VAE model integrates L1, L2, and perceptual loss functions to jointly optimize pixel-level accuracy and perceptual quality, while the LDM was trained using the Min-SNR-$\gamma$ strategy \cite{sup_hang2024efficientdiffusiontrainingminsnr} (see Supplementary Note 1 for a detailed description of VQ-VAE/LDM training strategy). To evaluate the reconstruction fidelity of thermal images and assess the realism of synthesized samples, we propose a modified Fréchet Inception Distance (FID) metric \cite{heusel2017gans} (see Supplementary Note 2 for a detailed description of the modified FID metric). For surface temperature reconstruction, the training objective combines L1 and perceptual losses to enhance local consistency and perceptual sharpness. 


\backmatter

\section*{Data Availability}
All data are available and can be accessed
through \href{https://zenodo.org/records/19222014}{https://zenodo.org/records/19222014}

\section*{Code Availability}
All codes are available and can be accessed
through \href{https://zenodo.org/records/19222014}{https://zenodo.org/records/19222014}

\section*{Acknowledgments}
We would like to thank the St. Agatha Fire Brigade Drone Unit and the Upper Austria Fire Brigade Headquarter for supporting the field experiments. 

\section*{Funding Information}
This research was funded by the State of Upper Austria and the Austrian Federal Ministry of Education, Science and Research via the LIT-Linz Institute of Technology under grant number LIT2019-8-SEE114. 

\section*{Author Contributions}

O.B. developed the concept, conceived and designed the algorithm and experiments.
M.Y. and L.B. implemented the algorithm and performed the experiments.
M.Y. , L.B. and O.B. analyzed the data, contributed materials, and wrote the paper.
G.C. and K.R. implemented and performed the field experiments.

\section*{Competing Interests}
The authors declare no competing interests.

\newpage

\section*{Supplementary Information}
\renewcommand{\figurename}{Supplementary Figure}
\setcounter{figure}{0}


\subsection*{Supplementary Figure 1 - Model Architecture for Forest Surface Temperature Generation}

\begin{figure}[H]
\centering
\includegraphics[width=1.0\textwidth]{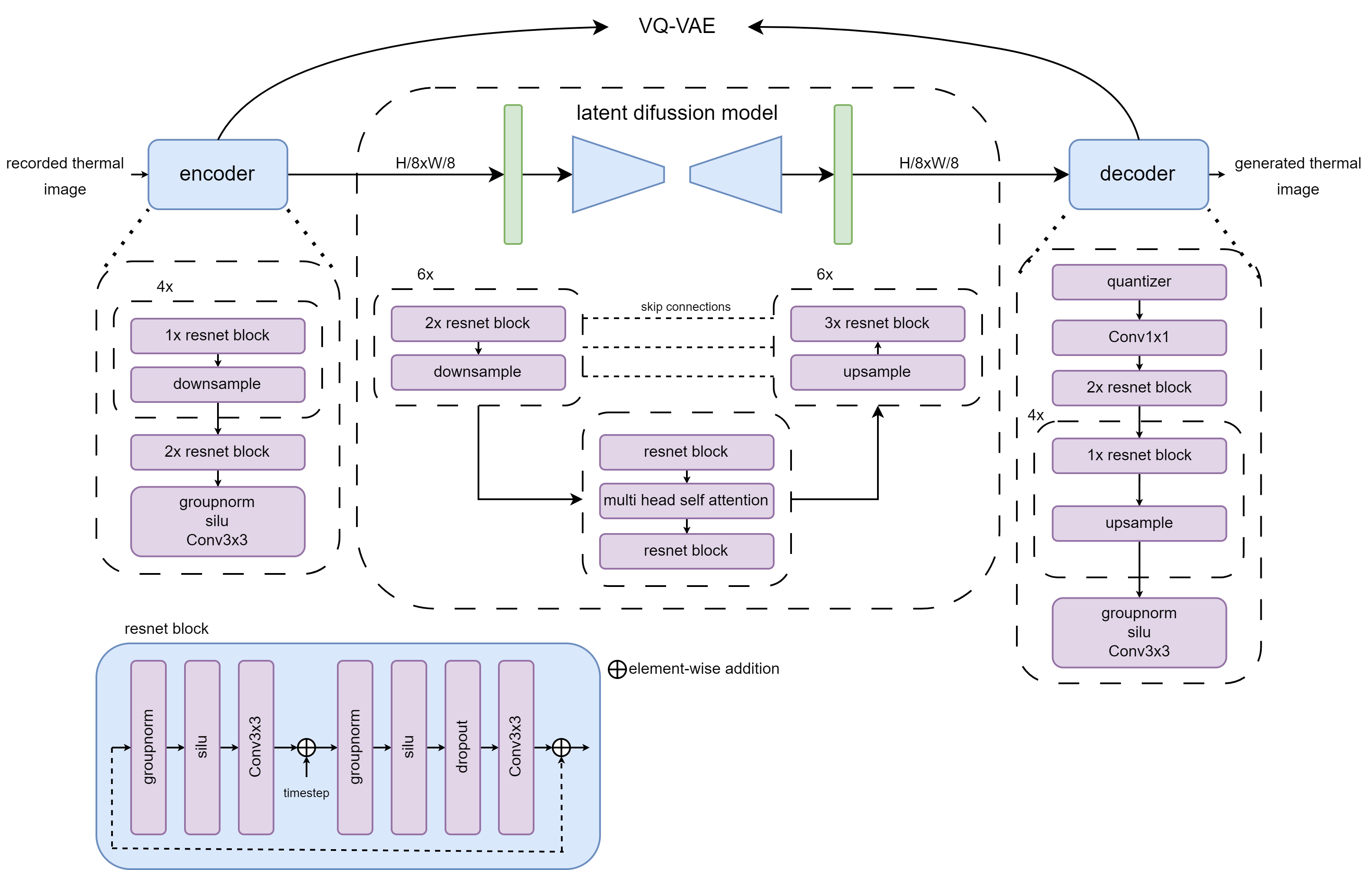}
\caption{Model architecture for forest surface temperature generation. Recorded thermal image is first encoded by a VQ-VAE into a latent space with reduced spatial dimensionality, which serves as the input to the LDM. The LDM operates within this compact representation to model the underlying data distribution, after which the latent features are decoded to reconstruct the corresponding high-resolution image. The output is generated thermal image with realistic spatial and thermal characteristics.  }\label{Fig:model_overview}
\end{figure}

\subsection*{Supplementary Figure 2 - Loss Function Evolution for Generating Surface Temperatures}
\begin{figure}[H]
     \centering
     \includegraphics[width=1.0\textwidth]{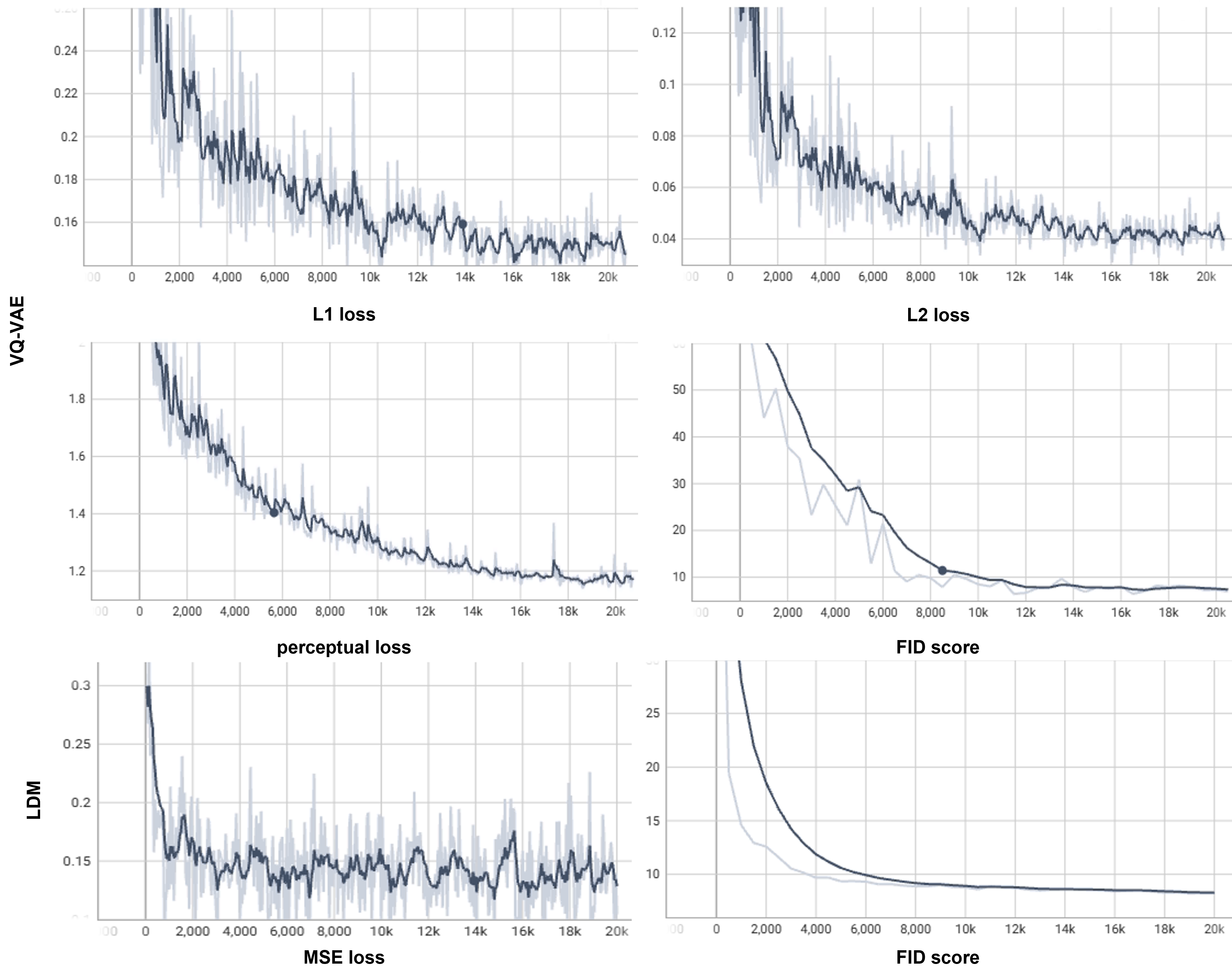}
     \caption{Evolution of the loss functions (here: L1, L2, perceptual loss, and Mean Squared Error, MSE) during the training of our VQ-VAE/LDM model, illustrating stable convergence behavior. The Fréchet Inception Distance (FID) score further demonstrates the high fidelity and perceptual quality of the generated thermal images.}
     \label{fig:VQ-VAE_LDM-analysis}
 \end{figure}

 \subsection*{Supplementary Figure 3 - Effect of different $\alpha$ values on The Transition Behavior}
\begin{figure}[H]
     \centering
     \includegraphics[height=5cm]{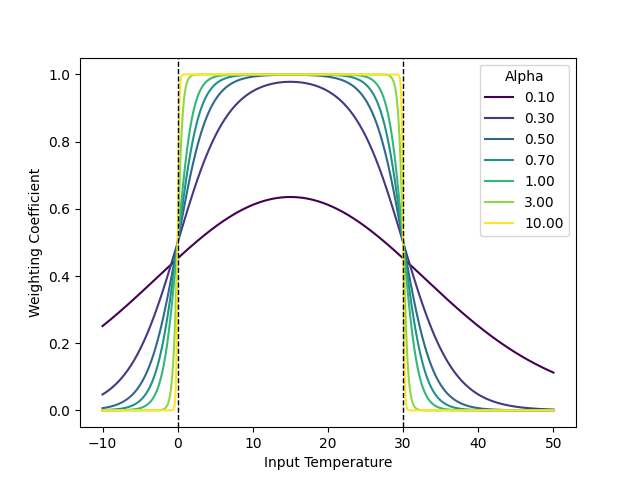}
     \caption{The sigmoid function $\psi(t)$ used in Eqn. 1 of the main manuscript to smoothen discontinuities at the boundaries of $T_{lower}$ and $T_{upper}$ (here, for example, $0^\circ C$ and $30^\circ C$) . The parameter $\alpha$ which controls smoothness of the temperature augmentation around lower and upper temperature thresholds. Empirically, we chose $\alpha$=0.5.}
     \label{fig:smooth_threshold}
 \end{figure}

 \subsection*{Supplementary Figure 4 - Temperature Augmentation}
\begin{figure}[H]
\centering
\includegraphics[width=1.0\textwidth]{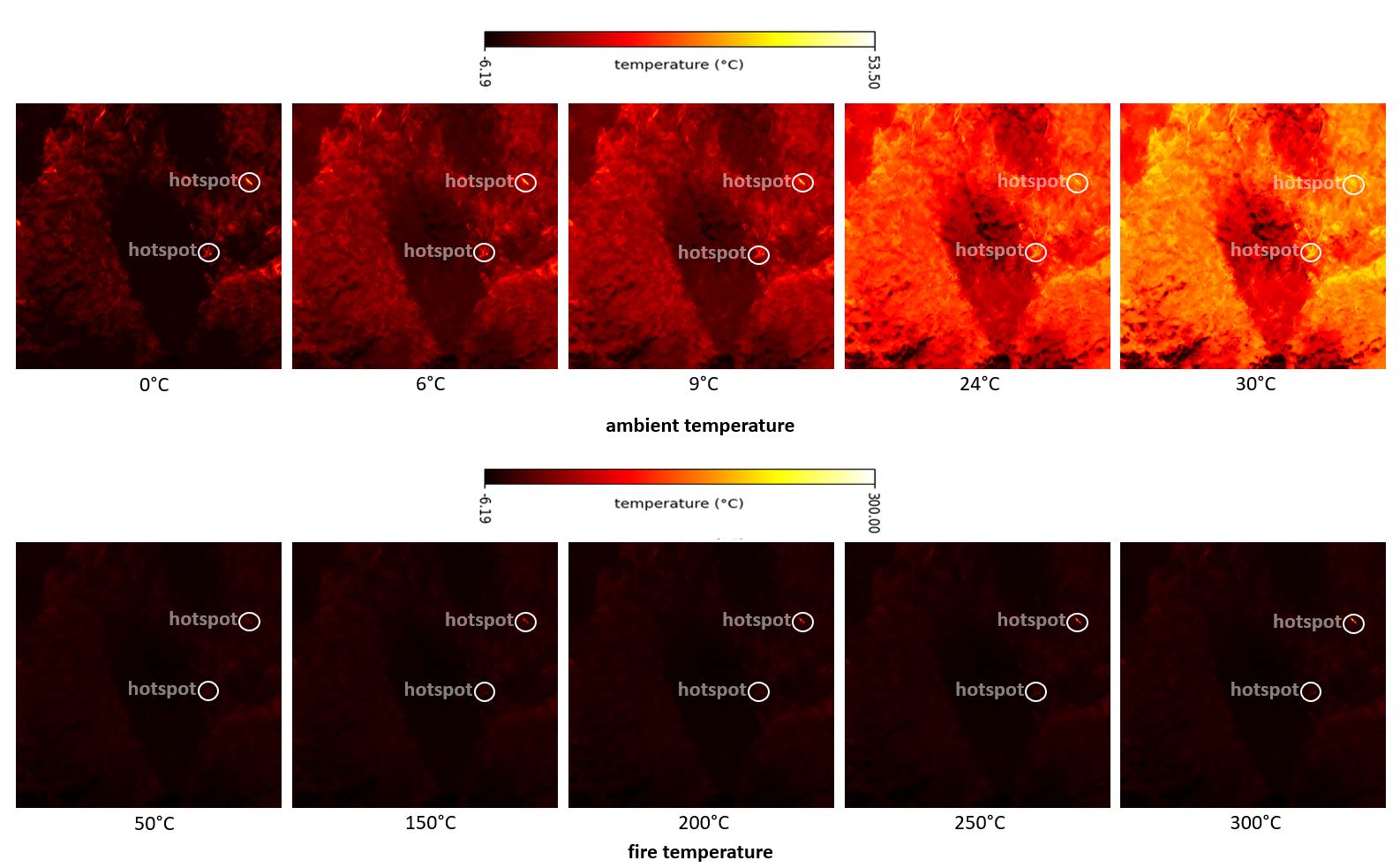}
\caption{Top row: An example of non-fire surface temperature augmentation at different ambient temperatures. Surface temperatures at 9$^\circ C$ ambient temperature (center) were generated by our VQ-VAE/LDM pipeline. These were then augmented to simulate lower (left) and higher (right) ambient conditions. The  temperatures of the visible fire hotspot is constant in all cases (53$^\circ C$ on average). Bottom row: For a constant ambient temperature of 9$^\circ C$, fire hotspot temperatures are scaled to a range of 50$^\circ C$ - 300$^\circ C$. Temperatures are color-coded according to the scales. Note that the bottom row's low-temperature pixels appear dark because we used a common and comparable color scale for all images. Only a few fire pixels are hot enough to appear bright.}\label{Fig:augmented_temp}
\end{figure}

\newpage

 \subsection*{Supplementary Table 1 - Parameters of Training and Test Datasets}
 
\begin{table}[h!]
\caption{Parameters of training and test datasets.}\label{Tab:DataParameters}
\begin{tabular}{@{}lll@{}}
\toprule
\textbf{parameter} & \textbf{training data} & \textbf{test data} \\
\midrule
forest density    & 220t/ha\footnotemark[1]   & 220-950t/ha\footnotemark[1] (steps: 70t/ha\footnotemark[1])   \\
ambient temperature    & 0-30$^\circ $C (steps: 1$^\circ $C) & 0-30$^\circ $C (steps: 3$^\circ $C)  \\
direct sunlight absorption\footnotemark[2]    & 0-15$^\circ $C (steps: 1$^\circ $C) & 0-15$^\circ $C (steps: 1$^\circ $C)   \\
solar azimuth \footnotemark[3]     & -90$^\circ$ - +90$^\circ$ (steps: continuous) & -90$^\circ$ - +90$^\circ$ (steps: continuous)    \\
\textbf{data size 1D SA (11 $\times$ 1)}  &  \textbf{5k $\times$ 11 = 55k images} & \textbf{2.4k $\times$ 11 = 26.4k images}    \\
\textbf{data size 2D SA (11 $\times$ 11)}  &  \textbf{5k $\times$ 120 = 600k images} & \textbf{2.4k $\times$ 120 = 288k images}    \\
\botrule
\end{tabular}
\footnotetext[1]{t/ha = trees per hectar}
\footnotetext[2]{As stated in \cite{kreye2018effects}, forest biomass can reach temperatures up to 15$^\circ C$ above ambient temperature, depending on direct sunlight absorption.}
\footnotetext[3]{Solar azimuth direction is fixed from east to west.}
\end{table}


 \subsection*{Supplementary Figure 5 - Visual Comparison of Simulated and Real Thermal Aerial Images of a Wildfire}
 
\begin{figure}[H]
\centering
\includegraphics[width=0.75\textwidth]{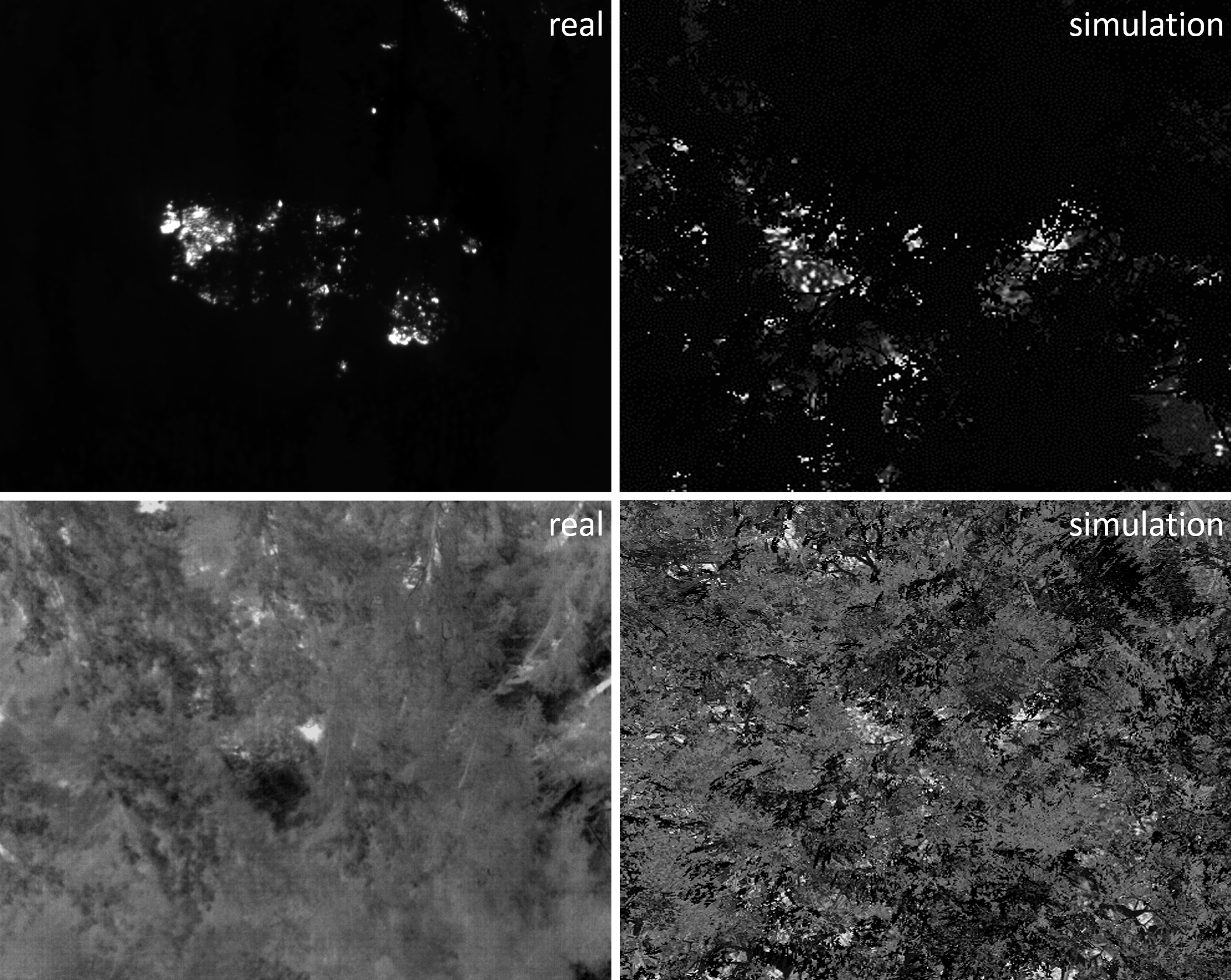}
\caption{A visual comparison of simulated and real thermal aerial images of a wildfire beneath the forest canopy, under low (top) and high (bottom) ambient temperatures. For a direct comparison, the simulated results are shown in grayscale to match the format of the real recordings. The simulations were parameterized using conditions from the real recordings—provided by the volunteer fire department—to closely match forest density, ambient temperature, solar azimuth, direct sunlight absorption, and fire temperatures.}\label{Fig:real_vs_simulation}
\end{figure}


 \subsection*{Supplementary Figure 6 - Model Architecture for Forest Surface Temperature Reconstruction}
\begin{figure}[H]
\centering
\includegraphics[width=1.0\textwidth]{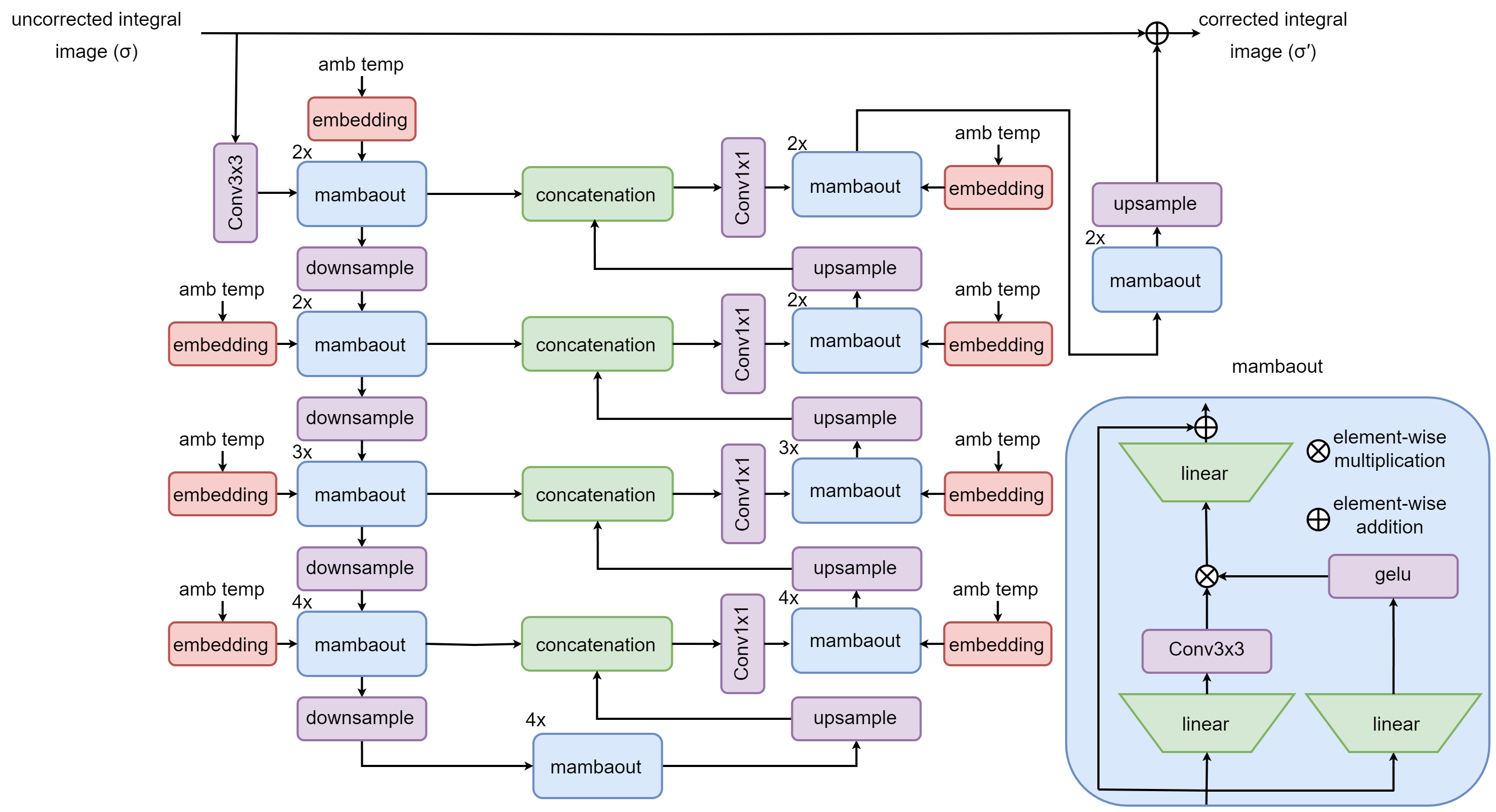}
\caption{Model architecture for forest surface temperature reconstruction. Uncorrected (AOS) integral image ($\sigma$) is first processed through an initial convolutional layer to extract low-level spatial features. These extracted features are subsequently fed into a U-Net architecture composed of MambaOut modules, which facilitate multi-scale feature extraction and reconstruction. The resulting feature maps are iteratively refined through multiple MambaOut processing stages. Finally, an upsampling layer reconstructs the corrected integral image ($\sigma'$).}\label{Fig:reconstructing_surface_temp}
\end{figure}

 \subsection*{Supplementary Figure 7 - Loss Function Evolution for Reconstructing Surface Temperatures}

\begin{figure}[H]
\centering
\includegraphics[width=1.0\textwidth]{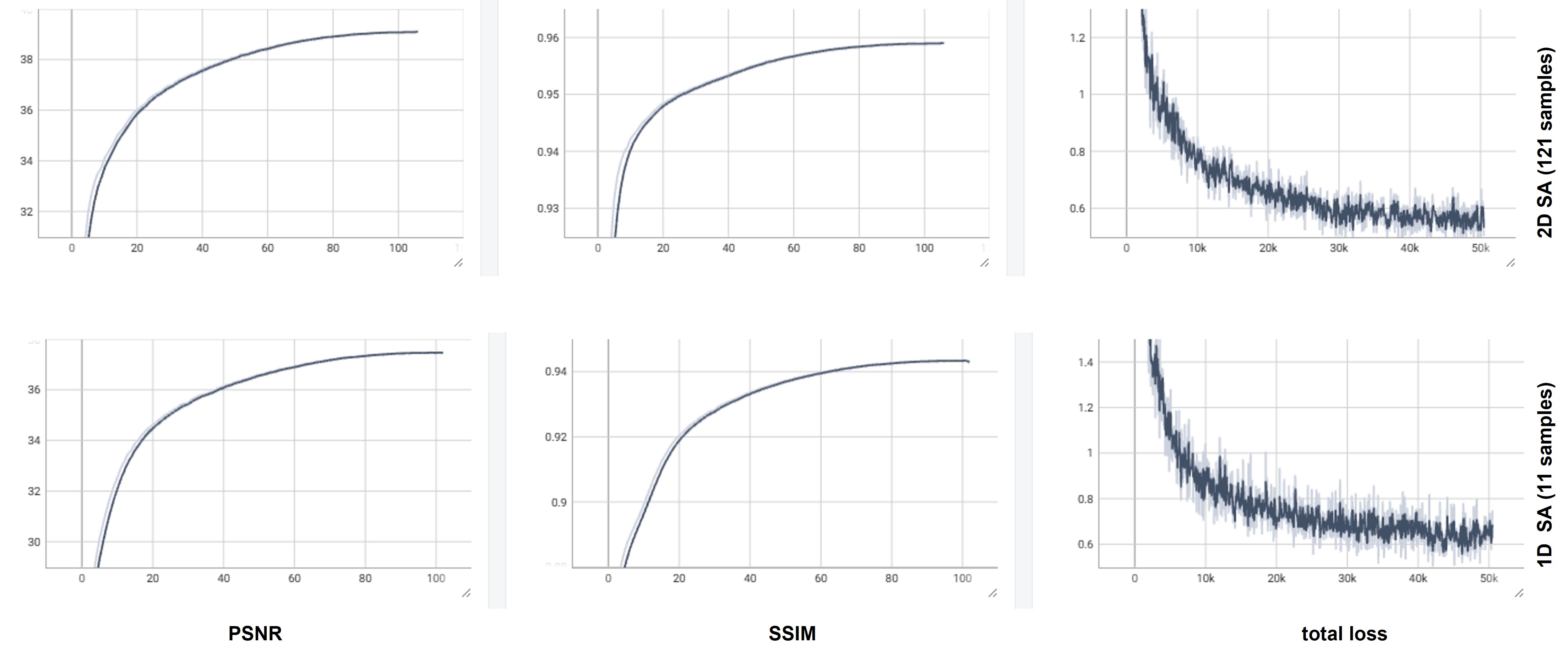}
\caption{The evolution of the loss functions (here: peak signal-to-noise ratio (PSNR), structural similarity index (SSIM), and total loss) during training of our VmambaIR framework with MambaOut blocks for 2D and 1D SAs demonstrate stable convergence.}\label{Fig:training_analysis}
\end{figure}


\subsection*{Supplementary Figure 8 - Field Experiment}

\begin{figure}[H]
\centering
\includegraphics[width=0.8\textwidth]{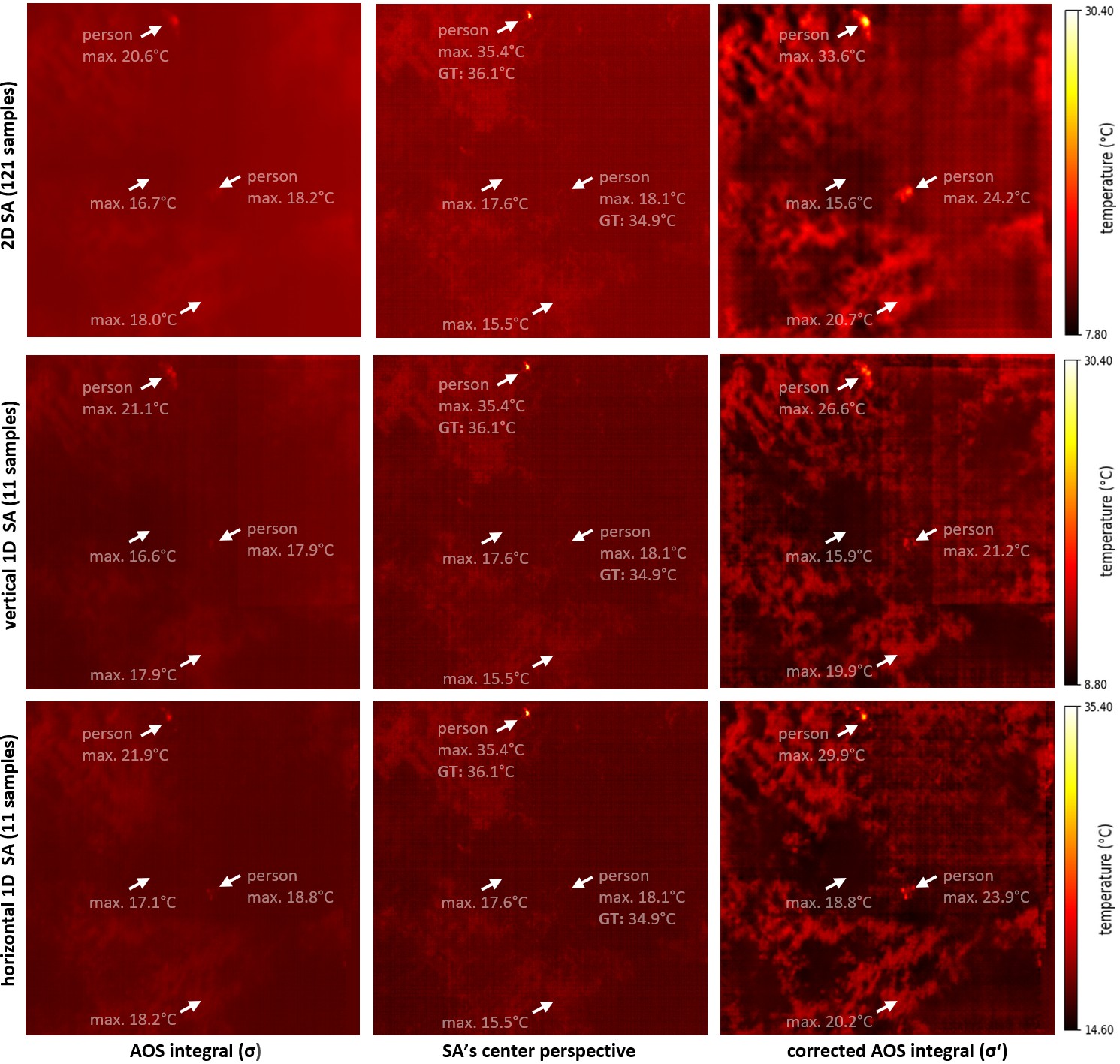}
\caption{Field experiment 3 involved the detection of two people beneath forest vegetation. The same flight path as in the previous two experiments (see Sect. 2.5 of the main manuscript) was conducted at the same location but at a different time (6 October 2025, drone flight: 5:53 PM–6:06 PM). The ambient temperature, estimated from the thermal recordings, was approximately $18^\circ C$. The results present three types of thermal imagery: AOS integral images (left column), which show maximal body and local surface temperatures derived from the integral images; regular thermal images from the synthetic aperture's (SA) center perspective waypoint (center column), which show maximal body and local surface temperatures (both ground truth (GT) and image-derived); and corrected AOS images (right column), which show maximal body and local surface temperatures from the corrected integral images. Sampling was performed using three approaches: a 2D SA (top row), a 1D SA as a vertical strip of the 2D SA (center row), and a 1D SA as a horizontal strip of the 2D SA (bottom row), with temperatures color-coded according to the provided scales. In contrast to regular thermal images, our approach successfully reconstructs the morphological characteristics, including the shape and extent, of the bodies. In contrast to uncorrected AOS integrals, it approximates the temperatures much better. Note that clothing significantly blocks body heat.}\label{Fig:SAR}
\end{figure}

\subsection*{Supplementary Note 1 - Training Strategy for Generating Surface Temperatures}

In our implementation, we employ a VQ-VAE with Lookup-Free Quantization (LFQ) and also evaluate vanilla Vector Quantization (VQ) and Finite Scalar Quantization (FSQ) \cite{sup_mentzer2023finite}. Although LFQ does not achieve the best reconstruction performance in terms of Fréchet Inception Distance (FID) \cite{heusel2017gans}, it yields the lowest generated FID score in the generative setting of our Latent Diffusion Model (LDM).

During training, the LDM is tasked with predicting velocity rather than the sample or noise, as done in previous diffusion approaches. Additional modifications include rescaling the noise schedule to enforce a zero terminal signal-to-noise (SNR) and adjusting the sampler to always start from the last timestep \cite{10484327}. These changes ensure that training and inference in the LDM remain congruent, leading to generated samples that more faithfully match the original distribution.

At inference time, the LDM takes random noise latents as input, and after 50 diffusion steps, the VQ-VAE decoder generates high-resolution thermal images. The LDM operates on a latent space of shape (1, 12, 64, 64), corresponding to a high-resolution input of (1, 1, 512, 512), representing an 8× spatial reduction. This compression substantially reduces the computational load during both training and inference, as the LDM must execute multiple denoising steps (typically 50) per sample.

We use a DDIM \cite{song2021denoising} noise scheduler with 1000 timesteps during training and 50 during inference, employing a scaled linear beta schedule specific to latent diffusion models with start and end values of 0.00085 and 0.012, respectively. An Exponential Moving Average (EMA) is maintained over the UNet weights, and gradient norms are clipped to a maximum of 1. Training is performed in mixed precision.

Instead of the conventional MSE loss between predicted and target velocities, we apply the Min-SNR-$\gamma$ strategy \cite{sup_hang2024efficientdiffusiontrainingminsnr} (with $\gamma$ = 5). This method adapts timestep loss weights based on clamped signal-to-noise ratios, effectively balancing gradient contributions across timesteps and improving the convergence speed of the LDM. As proposed in the original LDM paper \cite{9878449}, we scale the latent representations produced by the VQ-VAE to have unit variance and zero mean, using the batch statistics of the first training batch.

\subsection*{Supplementary Note 2 - Modified FID Metric}

We propose a modified FID metric. In the commonly used torch-fidelity package, FID computation resizes all input images to 299×299×3 before passing them through a pretrained Inception network designed for RGB images. In contrast, our implementation preserves the original 512×512 resolution, removes internal normalization and data type constraints, and replicates each thermal image across three channels to match the expected input shape of the pretrained Inception network.

\bibliography{sn-bibliography}

\end{document}